# Higher Order and Long-Range Synchronization Effects for Classification and Computing in Oscillator-Based Spiking Neural Networks

**Andrei Velichko[*], Vadim Putrolaynen, and Maksim Belyaev**

Petrozavodsk State University, Petrozavodsk, 185910, Russia
* Corresponding author: e-mail velichko@petrsu.ru

## Abstract

In the circuit of two thermally coupled $VO_2$ oscillators, we studied a higher order synchronization effect, which can be used in object classification techniques to increase the number of possible synchronous states of the oscillator system. We developed the phase-locking estimation method to determine the values of subharmonic ratio and synchronization effectiveness. In our experiment, the number of possible synchronous states of the oscillator system was twelve, and subharmonic ratio distributions were shaped as Arnold's tongues. In the model, the number of states may reach the maximum value of 150 at certain levels of coupling strength and noise. The long-range synchronization effect in a one-dimensional chain of oscillators occurs even at low values of synchronization effectiveness for intermediate links. We demonstrate a technique for storing and recognizing vector images, which can used for reservoir computing. In addition, we present the implementation of analog operation of multiplication, the synchronization based logic for binary computations, and the possibility to develop the interface between spike neural network and a computer. Based on the universal physical effects, the high order synchronization can be applied to any spiking oscillators with any coupling type, enhancing the practical value of the presented results to expand spike neural network capabilities.

## 1. Introduction

A real biological neuron is a complex biochemical system that operates with a continuously incoming multichannel stream of voltage pulses - spikes [1]. Sequences of spikes arrive at the synapses of the neuron, while the neuron itself also generates a sequence of spikes that propagate along neuron's output process - the axon. In the framework of cybernetics and computer science, the practical question is - what features of the biological neurons can help the neural networks to solve practical problems, such as pattern recognition, adaptive control problems [2–5], and synchronization over long distances [6–8]? A number of the information coding methods for spiking neural networks (SNN) has been developed [9,10]: rate coding, rank coding, time to first spike, latency coding, phase coding, and synchronous state coding [11–13]. The latter method is used for oscillatory neural networks (ONNs), which are a type of SNN, where the processes of neurons synchronization are the main effects that control the operation of the network. The study of ONN is a promising research direction, as cognitive functions of the brain are associated with the effect of synchronization of various groups of neurons within certain time windows [6]. Substantial progress has been achieved in the understanding of the synchronization and de-synchronization processes in the networks of interacting oscillators [14]. The results of these studies have been used to solve a wide variety of applied tasks, for instance, pattern recognition [11,15–17], information classification [12,13,18–20] and computational operations [21–24]. Various oscillator circuits have been suggested for these tasks based on different physical phenomena: oxide-based oscillators [25–28], nanomechanical oscillators [24,29], spin-torque nano-oscillators [21,30–33], laser oscillators [34] and superconducting oscillators [35]. The current study adds new insights to the field of synchronization in SNN, contributing to solving problems of classification and computing.

Although the effect of oscillators' synchronization had a long research history and was observed as far back as in the 17th century by Dutch physicist Christiaan Huygens, the new aspects of this phenomenon continue to be discovered and find experimental confirmation in various sciences. The effect of the formation of chimeric states was discovered, when some of the oscillators are synchronized and some oscillators are not synchronized [36]. The possibility of using higher-order synchronization and its more complex version, chimeric synchronization, to create neural networks was predicted [11–13]. Chimeric synchronization [13], also known as "hopping mode" [37], when synchronization patterns alternate in time, is the most complex variation of the synchronization effect, and the study of its role in biological [38] and physical processes requires additional efforts. In this regard, the study of the synchronization effects of two oscillators and the search for experimental objects for synchronization observation and research are the important task.

Among non-silicon materials, vanadium dioxide-based switches can be highlighted, and neurons based on them are of interest to many researchers [25,39–41]. The reason for this interest lies in the possibility of a simple implementation of oscillators on $VO_2$ switches, which have the shape of current peaks similar to spikes of real neurons. In the oscillatory $VO_2$ system, internal noise is always present due to the instability of the threshold characteristics of the switching channel [42,43], and the creation of a stochastic $VO_2$ neuron removed this drawback [44,45]. The introduction of noise into spike models is justified, since there are a large number of noise sources in a real neuron [46,47], for example, release probability of input spikes at synapses [48].

In the current study, we used two oscillators based on $VO_2$ switches as experimental and model objects. The mechanism of oscillators' interaction was implemented through thermal coupling, as described in details in [49,50]. The advantage of the thermal coupling with respect to capacitive or resistive coupling is the complete electrical isolation of the oscillators from each other, both in alternating and in direct current [51]. It prevents the oscillators from shifting the operating points and makes them more stable, when the control parameters are varied.

The study of the synchronization effect of two $VO_2$ oscillators to create an effective oscillatory neural network is the main objective of the current article. Spike oscillations have a wide, linear spectrum of frequencies. During thermal interaction of oscillators, individual spikes in each circuit appear in phase, and it leads to a phase-locking effect. These features cause the strong effect of higher order synchronization [14], which is expressed in the synchronization of oscillations on individual subharmonics of the spectrum. However, the use of high order synchronization for classification and pattern recognition in SNN has been insufficiently studied.

The problem of determining the synchronization state is intensively discussed in the academic literature [14]. Lowet et al. [52] compare two main approaches - spectral coherence and phase-locking value approaches, and the authors conclude that phase-locking value approach is more preferable, especially, in the presence of significant noise and partially (intermittent) synchronized state. We present an alternative efficient algorithm [12] that determines the presence of a high-order synchronization effect and enables its quantitative assessment. The algorithm description reveals the relationship between spectral coherence and phase-locking value approaches and reviews the advantages of the latter. We developed this algorithm for the estimation of chimeric synchronization (hopping mode) earlier [13]. In the current article, we discuss the application of the algorithm for studying the effect of long-range synchronization.

We estimate the SNN efficiency by the value of the classification capacity parameter, which is determined by the number of possible states the oscillator system can take. The classification problem is closely related to the pattern recognition problem. Vodenicarevic et al. [18] proposed a circuit based on 4-coupled oscillators, which is resistant to noise, parameter variations and nonlinearity of the oscillators, and has about nine stable states. In this study, we propose an alternative approach to solving this problem, which would lead to an increase in the classification capacity of oscillatory circuits with a smaller number of oscillators.

Systems of coupled oscillators find applications in non-Boolean calculations (analog logic) [16,17,21,53,54], where the input parameters are specified through the physical characteristics of the circuit, such as oscillator frequencies, phase differences, or communication forces between them. The output of such system is the physical characteristics of the circuit in its final stable state. The calculations in analog oscillator systems require parallel processing of information, which amount depends on the number of oscillators included in the network. The idea to apply phase-based logic to oscillator operations was suggested by John von Neumann [55]. Roychowdhury [56] demonstrated the advantages of this approach in comparison with level-based CMOS computing, such as high noise resistance and low energy consumption. In this study, we develop the principles of calculations based on the effect of high order synchronization and suggest our version of the synchronization-based logic.

## 2. Methods

### 2.1 Experimental techniques

In our previous study [49], we described the process of $VO_2$ films deposition and $VO_2$ films-based planar structures. Coupled structures (Fig. 1a) were formed with the distances of $d$~12 μm and $d$~21 μm to ensure, respectively, strong and weak thermal couplings between them [49,50]. The maximum temperature induced from a neighboring switch was $\Delta T^p \sim 0.7$ K at strong coupling and $\Delta T^p \sim 0.2$ K at weak coupling. The experimental I-V characteristic of an individual switch, which has a classic S-shape, is shown in Fig. 1b. The basic parameters of I-V characteristic are threshold currents ($I_{th}$, $I_h$), threshold voltages ($U_h$, $U_{th}$), cutoff voltage ($U_{cf}$) and dynamic resistances at high state and low state ($R_{off}$ and $R_{on}$). The detailed description of the structure characteristics and thermal coupling is given in our previous publications [57–59].

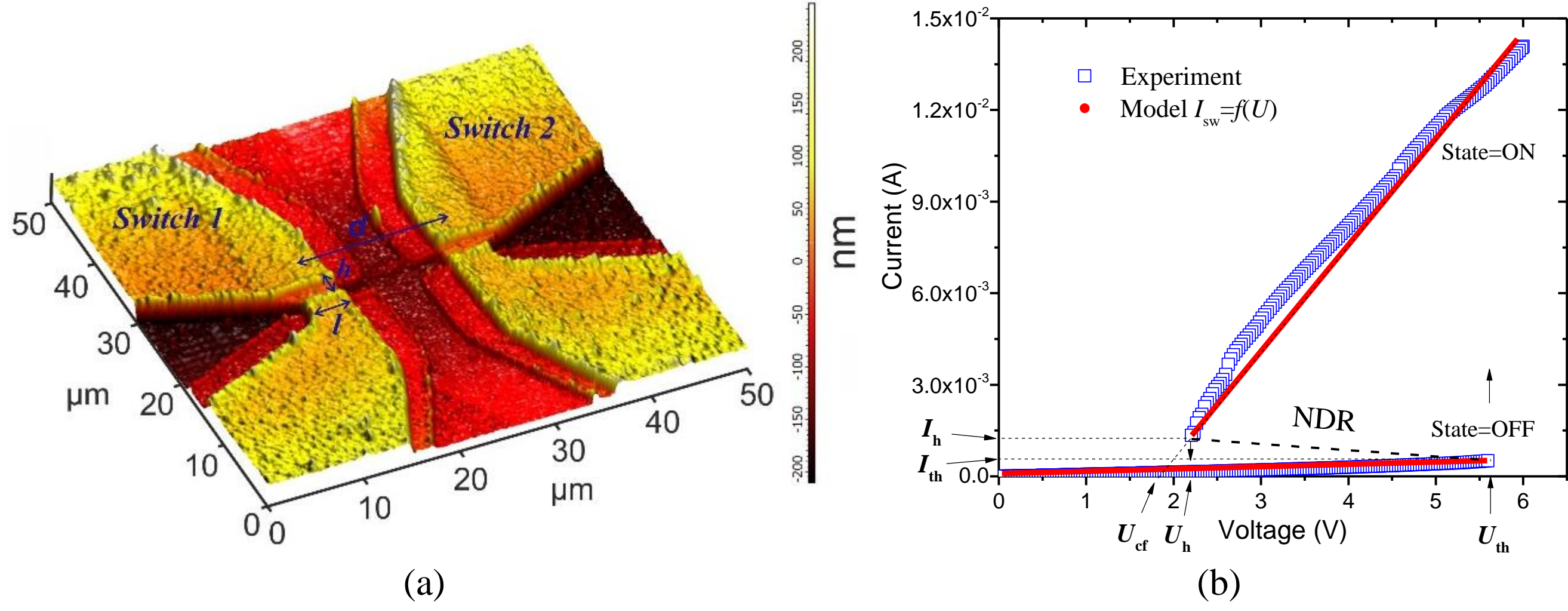

(a) (b)

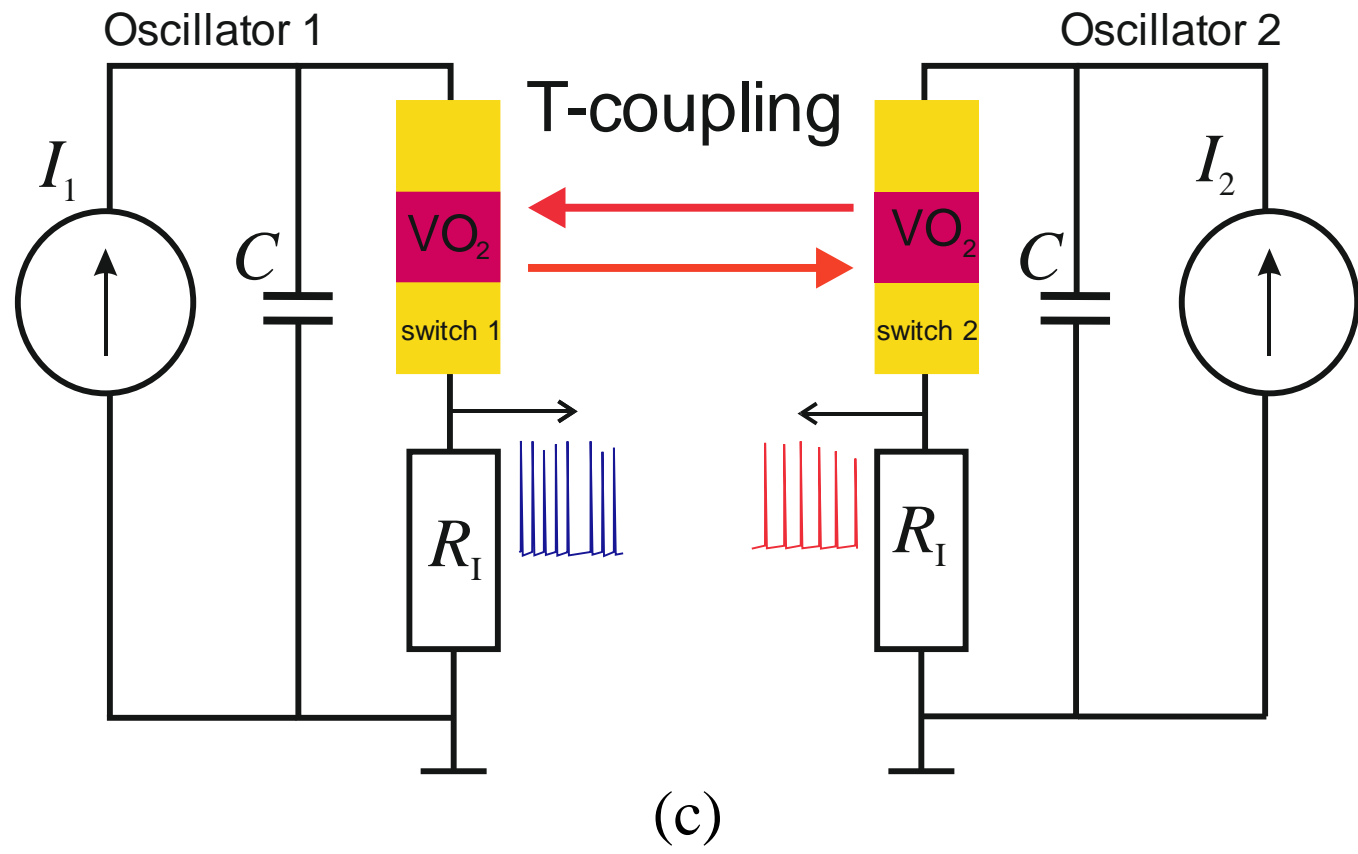

**Figure 1** [Color online] AFM image of coupled planar $VO_2$-structures (a), typical experimental I-V characteristic of a separate switch (Experiment curve) and its model curve (Model curve) (b). Experimental circuit of two oscillators with thermal coupling (c). Length *l* and gap *h* of the switch inter-electrode space are ~ 3-4 µm and 2.5 µm, respectively.

Experimental circuit of two oscillators is presented in Fig. 1c. The circuit of each oscillator has a current source $I_{1(2)}$, capacity $C$ (100 nF) and limiting current resistor $R_I$ (250 Ω) connected in series with $VO_2$-switch. The current of output signal is registered on the resistor $R_I$. Oscillations in a single circuit are studied in details in the academic literature [16,25,43,51,60], and such oscillations are observed, when the current of power supply is within the range of $I_{th} < I_{1(2)} < I_h$. This condition ensures that the circuit's operating point is in negative differential resistance area (NDR) (Fig. 1b). In the presence of a strong connection between the two circuits [50], the lower boundaries of the current range of oscillation existence expand, because the threshold voltage $U_{th}$ decreases due to mutual pulsed Joule heating and causes a decrease in $I_{th}$.

The study of oscillation dynamics was performed with a four-channel oscilloscope Picoscope 5442B, having maximum sampling rate of 125 MS/sec in 14-bit mode. A two-channel sourcemeter Keythley 2636A was used for the DC *I-V* characteristic measurements (sweeping rate 1V/s) and as the current source for oscillatory circuit.

A surface morphology characterization was conducted using AFM NTEGRA Prima in non-contact mode.

### 2.2 Model of an oscillator and thermal coupling

A model diagram for calculating current (voltage) oscillograms in coupled oscillatory circuits is presented in Fig. 2b. The diagram repeats the experimental circuit, except for the absence of a current resistor and the presence of a controlled noise generator $U_n$. Connecting $U_n$ allows the simulation of real internal circuit noise that occurs due to fluctuations of the current channel in the switching structure [42,61]. In addition, the noise generator allows the analysis of the influence of the external noise amplitude on the characteristics of the synchronization effect.

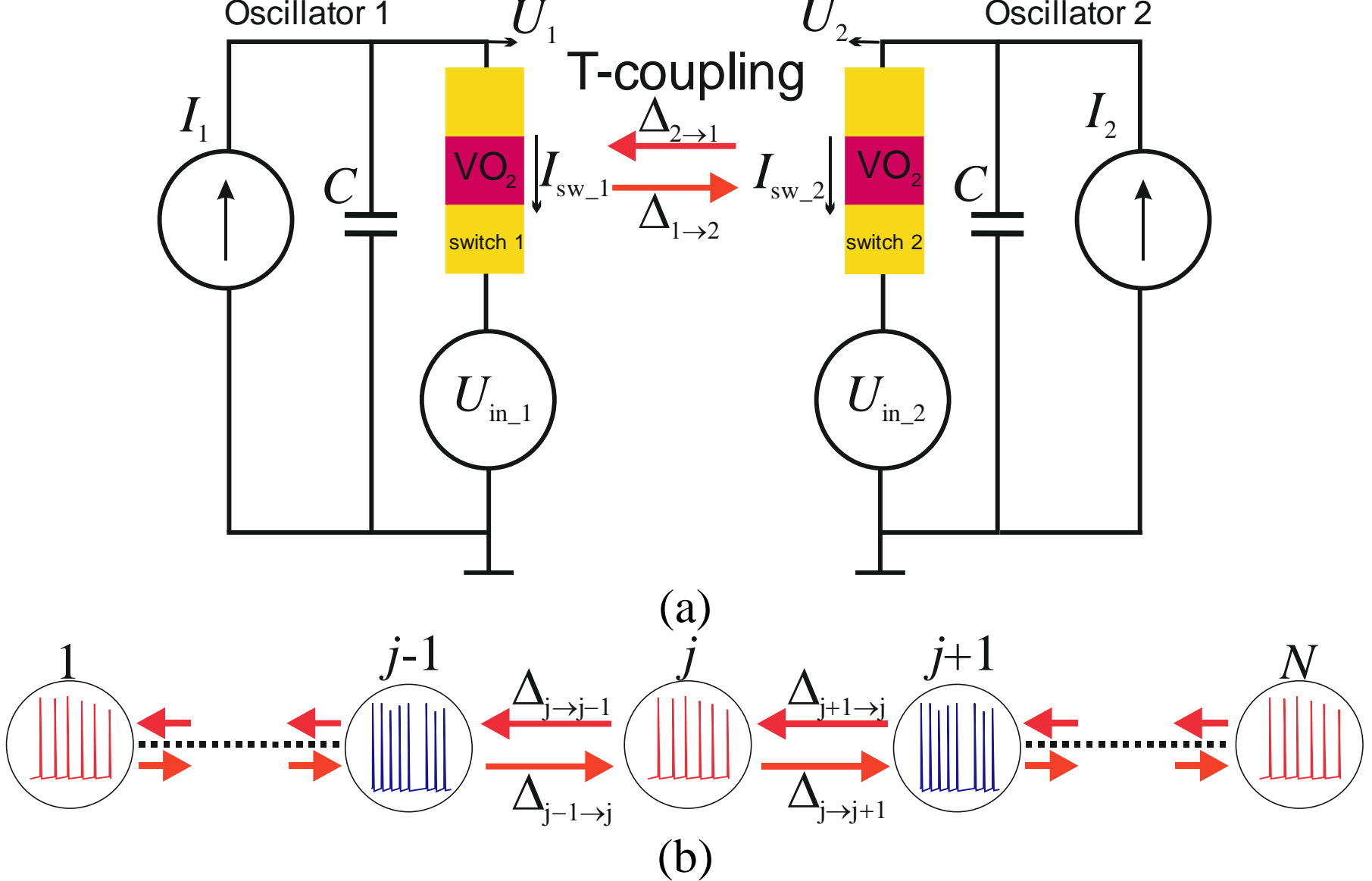

**Figure 2** [Color online] Model circuit of two oscillators with thermal coupling (a) and *N*- oscillators chain (b).

To calculate the model circuit, a simple system of differential equations was solved:

$$\begin{cases} C\dfrac{dU_1(t)}{dt} = I_1 - I_{\text{sw_1}}(t), \text{ where } I_{\text{sw_1}}(t) = f(U_1(t) - U_\text{n}(t)) \\ C\dfrac{dU_2(t)}{dt} = I_2 - I_{\text{sw_2}}(t), \text{ where } I_{\text{sw_2}}(t) = f(U_2(t) - U_\text{n}(t)) \end{cases}, \tag{1}$$

where $U_{1(2)}(t)$ is voltage at capacitors, $I_{1(2)}$ is power current of constant value, $I_{\text{sw_1(2)}}(t)$ is current via switches and ($U_{1(2)}$ $(t)$ - $U_\text{n}(t)$) is voltage at switches at the corresponding circuits, function $f(U)$ is determined by linear I-V characteristic of structure (2-3). In this case, the parameters of I-V characteristic depend on the state of the neighbor switch (4). System of Equations (1) was numerically solved with a uniform time step $\Delta t = 10^{-5}$ s by the implicit Euler method. The discrete noise $U_\text{n}$ $(t)$ was generated by the algorithm $U_\text{n}(t) = U_{\text{n0}} \cdot \text{randn}(t)$, where $U_{\text{n0}}$ is noise amplitude and randn(t) is normal random numbers generator with zero mean and dispersion equal to 1 [50].

To form the function $f(U)$, both branches of I-V characteristic (Fig. 1b, Model curve) were approximated by straight-line segments with dynamic resistances $R_\text{off}$ and $R_\text{on}$:

$$f(U) = \begin{cases} \dfrac{U}{R_{off}}, & \text{if } State = \text{OFF} \\ \dfrac{(U - U_\text{cf})}{R_{on}}, & \text{if } State = \text{ON} \end{cases}, \tag{2}$$

where "*State*" denotes the switch state (OFF- high, ON – low).

Transitions from one state into another for two switches ($State_1$ and $State_2$) were modeled according to the following algorithm:

$$\begin{aligned} State_1 &= \begin{cases} \text{OFF}, & \text{if } (State_1 = \text{ON}) \text{ and } (U_1 < U_\text{h}) \\ \text{ON}, & \text{if } (State_1 = \text{OFF}) \text{ and } (U_1 > U_{\text{th_1}}) \end{cases} \\ State_2 &= \begin{cases} \text{OFF}, & \text{if } (State_2 = \text{ON}) \text{ and } (U_2 < U_\text{h}) \\ \text{ON}, & \text{if } (State_2 = \text{OFF}) \text{ and } (U_2 > U_{\text{th_2}}) \end{cases} \end{aligned}, \tag{3}$$

where $U_{\text{th_1}}$ and $U_{\text{th_2}}$ are threshold turn-on voltages of switches according to the algorithm:

$$
\begin{aligned}
U_{\mathrm{th_1}} &= \begin{cases} U_{\mathrm{th}} - \Delta_{2\to 1}, & \text{if } State_2 = \mathrm{ON} \\ U_{\mathrm{th}}, & \text{if } State_2 = \mathrm{OFF} \end{cases}, \\
U_{\mathrm{th_2}} &= \begin{cases} U_{\mathrm{th}} - \Delta_{1\to 2}, & \text{if } State_1 = \mathrm{ON} \\ U_{\mathrm{th}}, & \text{if } State_1 = \mathrm{OFF} \end{cases}
\end{aligned} \tag{4}
$$

where $\Delta_{1\to 2}$ and $\Delta_{2\to 1}$ (coupling strength) determine the degree of influence of one switch on another switch, with thermal coupling (Fig. 2a). For example, the value $\Delta_{1\to 2}$ reflects how much the value of the threshold voltage $U_{\mathrm{th_2}}$ (switch 2) changes, when the switch 1 switches to a low-impedance state. The interaction of oscillators occurs because of mutual influence on the threshold characteristics. The similar coupling types are found in the literature [14,62]. However, in our circuit, the electrical connection between the oscillators is absent; it greatly simplifies the system (1) and helps to simulate a network of a large number of coupled oscillators.

Studying the effect of long-range synchronization for the chain of $N$ oscillators, we used the model presented in Fig. 2b, where the interactions of only neighboring oscillators were taken into account. The following algorithm determined $U_{\mathrm{th_j}}$ of each oscillator inside the chain:

$$
U_{\mathrm{th_j}} = \begin{cases} U_{\mathrm{th_j}} - \Delta_{\mathrm{j\text{-}1}\to \mathrm{j}}, & \text{if } (State_{\mathrm{j\text{-}1}} = \mathrm{ON}) \text{ and } (State_{\mathrm{j+1}} = \mathrm{OFF}) \\ U_{\mathrm{th_j}} - \Delta_{\mathrm{j+1}\to \mathrm{j}}, & \text{if } (State_{\mathrm{j\text{-}1}} = \mathrm{OFF}) \text{ and } (State_{\mathrm{j+1}} = \mathrm{ON}) \\ U_{\mathrm{th_j}} - \Delta_{\mathrm{j\text{-}1}\to \mathrm{j}} - \Delta_{\mathrm{j+1}\to \mathrm{j}}, & \text{if } (State_{\mathrm{j\text{-}1}} = \mathrm{ON}) \text{ and } (State_{\mathrm{j+1}} = \mathrm{ON}) \\ U_{\mathrm{th_j}}, & \text{if } (State_{\mathrm{j\text{-}1}} = \mathrm{OFF}) \text{ and } (State_{\mathrm{j+1}} = \mathrm{OFF}) \end{cases} . \tag{5}
$$

At the chain edges for ($j$=1, $j$=$N$), Equation (4) was used. Equation (5) describes the decrease in $U_{\mathrm{th}}$ because of additional heating of the channel by $\Delta T^{\mathrm{p}}$, caused by the heating of neighboring switches at the moment of switching on. The highest decrease in $U_{\mathrm{th}}$ is observed, when both neighboring switches are turned on ($\mathrm{State}_{\mathrm{j\text{-}1}}$=ON and $\mathrm{State}_{\mathrm{j+1}}$=ON). In the simplified model, we assume that the thermal effect is reflected only in the $U_{\mathrm{th}}$ change, and the Δ value does not depend on the duration of the low-resistance state. The results of numerical simulations [49,57] demonstrated that $\Delta T^{\mathrm{p}}$ has a low value of time constant for reaching the stationary level, corresponding to the duration of the front of the current pulse (~ 3 ns, for our configuration). An increase in temperature has a stronger effect on $U_{\mathrm{th}}$ than on $U_{\mathrm{h}}$ so, our approximations are correct.

To simulate the experimental two-oscillator circuit (Fig. 1c) with the distance between the switches $d$ ~ 21 μm, we used the following I–V characteristics: $I_{\mathrm{th_1}}$= 390 μA, $I_{\mathrm{h_1}}$= 1100 μA, $U_{\mathrm{th_1}}$=5 V, $U_{\mathrm{h_1}}$=1.5 V $U_{\mathrm{cf_1}}$=0.8V, $R_{\mathrm{off_1}}$=13 kΩ, $R_{\mathrm{on_1}}$= 620Ω, $I_{\mathrm{th_2}}$= 370 μA, $I_{\mathrm{h_2}}$= 1330 μA, $U_{\mathrm{th_2}}$=5.4 V, $U_{\mathrm{h_2}}$=1.7 V, $U_{\mathrm{cf_2}}$=1 V, $R_{\mathrm{off_2}}$= 14.5 kΩ, $R_{\mathrm{on_2}}$= 530Ω. For switches with d ~ 12 μm and the subsequent model calculations, we used the following: $I_{\mathrm{th_1}}$= 550 μA, $I_{\mathrm{h_1}}$= 1100 μA, $U_{\mathrm{th_1}}$=5 V, $U_{\mathrm{h_1}}$=1.5 V, $U_{\mathrm{cf_1}}$=0.8 V, $R_{\mathrm{off_1}}$=9.1 kΩ, $R_{\mathrm{on_1}}$= 620Ω, $I_{\mathrm{th_2}}$= 450 μA, $I_{\mathrm{h_2}}$= 1330 μA, $U_{\mathrm{th_2}}$=5.4 V, $U_{\mathrm{h_2}}$=1.7 V, $U_{\mathrm{cf_2}}$=1 V, $R_{\mathrm{off_2}}$= 12 kΩ, $R_{\mathrm{on_2}}$= 530Ω. To simulate long-range synchronization, the switches were considered identical with the parameters corresponding to switch No. 1 at $d$ ~ 12 μm.

In all calculations, we assumed that $\Delta_{1\to 2}$=$\Delta_{2\to 1}$=Δ ($\Delta_{\mathrm{j}\pm 1\to \mathrm{j}}$=$\Delta_{\mathrm{j}}$=Δ), in other words, the mutual influence of the oscillators is equal. We can assume, when $\Delta_{1\to 2}\neq\Delta_{2\to 1}$, for example, $\Delta_{1\to 2}$=0 $\Delta_{2\to 1}\neq 0$, it is possible to simulate the unidirectional action of one oscillator on another oscillator, as it often happens in biological systems [63]. It can be a subject of further studies of the presented model.

### 2.3 Phase-locking estimation method

To evaluate synchronization, we introduced the synchronization efficiency parameter $\eta$ and subharmonic ratio parameter *SHR* in our previous studies [12,50], where we described the calculation methods for the parameters. The meaning of *SHR* can be interpreted as the ratio of the number of the signal subharmonics ($k_1$, $k_2$ see Fig. 3a) at the synchronization frequency $F_S$, expressed by the equation:

$$SHR = \frac{k_2}{k_1}. \tag{6}$$

Equation (6) uses spectral coherence approaches to describe synchronization of the higher order $k_2$ : $k_1$, with the following relation:

$$F_S = k_1 \cdot F_1^0 = k_2 \cdot F_2^0, \tag{7}$$

where $F_1^0$, $F_2^0$ are frequencies of the fundamental harmonics.

Alternatively, *SHR* value can be estimated by the phase-locking method:

$$SHR = \frac{M_1}{M_2}, \tag{8}$$

where $M_1$ and $M_2$ are the most probable number of signal periods that fit on the synchronization period $T^i_s$ of two oscillators (see Fig. 3b, where $i$ is the number of periods $T_s$). The examples of the spectra and oscillogram of the currents $I_{sw_1(2)}$ in Figure 3 correspond to a strong coupling at $I_1$=580 μA, where $T^i_s$ is defined as the time between two synchronous current peaks. With the chaotic behavior of the system, the synchronization periods differ and have a spread in the values of $T^i_s \neq T^{i+1}_s$ and in the values of $M$; therefore, the most probable values of $M_1$ and $M_2$ are used in Equation (8).

According to Equations (6, 8), *SHR* varies in a discrete way and can be represented as a simple fraction (1/1, 1/2, 2/3, etc.), because ($k_1$, $k_2$) and ($M_1$, $M_2$) belong to the set of integers. Various values of synchronizations *SHR* may occur within one oscillogram. To determine which *SHR* value prevails, it is necessary to find the occurrence probabilities $P(M_1{:}M_2)$ for each pair ($M_1{:}M_2$) that is present in the whole oscillogram and to select the pair with the maximum value of $P = P_{\max}(M_1{:}M_2)$. Then, the final value of *SHR* will be written as [12]:

$$SHR = M_1 : M_2, \text{if } P = P_{\max}(M_1 : M_2). \tag{9}$$

To find the probabilities $P(M_1{:}M_2)$, we can count how many times $NP(M_1{:}M_2)$ the given pair appeared within the whole oscillogram of the oscillator 2, multiply by the number of periods in it ($M_2$) and divide by the total number of all oscillations periods in the given signal ($N_2$). Thus, for $P(M_1{:}M_2)$ we obtain:

$$P(M_1 : M_2) = 100\% \cdot NP(M_1 : M_2) \cdot M_2 / N_2, \tag{10}$$

where $N_2$ is the total number of periods in the oscillogram of oscillator 2.

For *SHR*, the parameter of synchronization effectiveness $\eta$ is defined as the maximum probability $P_{max}(M_1{:}M_2)$:

$$\eta = P_{\max}(M_1 : M_2). \tag{11}$$

Finally, we compared the values of $\eta$ with the selected threshold value of $\eta_{th}$ = 90%. If $\eta \geq \eta_{th}$, the oscillations were considered synchronized, otherwise, oscillations were considered chaotic. The results would depend on the choice of $\eta_{th}$.

An additional restriction was the maximum values of $M_1$ and $M_2$. If the values did not satisfy the condition ($M_1$ or $M_2$) ≤ 20, then the oscillations were considered not synchronized. For example, we considered *SHR* = 50/48 not synchronized. This restriction applies if the model implies a reduction in the noise level to a low value, tending to zero $U_{n0} \rightarrow 0$. In real experimental systems with a high noise level, we did not achieve synchronization at high order harmonics. This condition is closely related to the determination of the maximum classification capacity of the $W_C$ system.

Equations 6–11 reflect the methodology for evaluating the *SHR* and $\eta$ synchronization metrics for a network of two oscillators. In the general case, when there are $N$ oscillators in the system, the synchronization metrics between any two oscillators with numbers $i$ and $j$ are calculated similarly and are denoted as $SHR_{i,j}$ and $\eta_{i,j}$.

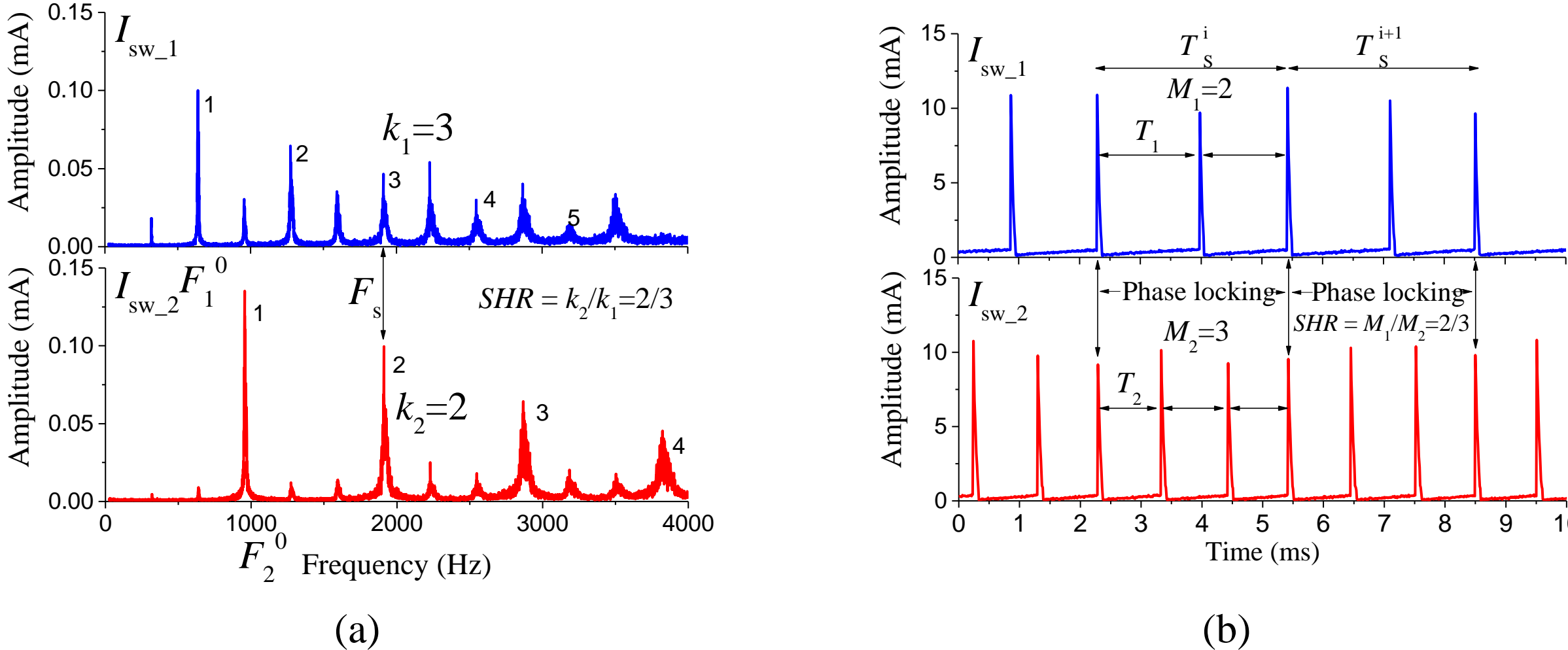

**Figure 3** [Color online] View of the experimental spectra (a) and current oscillograms (b) of two coupled oscillators, reflecting the algorithm for calculating the phase synchronization parameters (experiment conditions correspond to strong coupling at $I_1$ = 580 μA).

The phase locking method (Equations 8-11) is more efficient in the *SHR* search than the spectral approach (Equations 6, 7). First, the algorithm is resistant to period fluctuations, compared to the spectrum method. In the spectrum method, the spectrum line broadens, and the synchronization frequency $F_S$ and the amplitude cannot be determined clearly, as the amplitude decreases with an increase in the number of the subharmonic. Second, the determination of the synchronization efficiency $\eta$ and the condition ($\eta \geq \eta_{th}$) enable a clear separation of the synchronous state of the system from the non-synchronous (chaotic) state. Third, in contrast to the spectral method, where the resource-intensive fast Fourier transform is applied, the phase locking method uses only comparison and addition operations. It significantly speeds up and simplifies the calculations.

The presented phase-locking estimation method can be applied not only to current oscillograms, but equally to voltage oscillograms on switches. The main point is to search for sharp edges, which are synchronous in time and appear, when the structure is turned on.

The method for *SHR* calculation (Equations 8-11) can be developed further, for example, for the cases, when two types of synchronization with different *SHR* values simultaneously exist in the system. It happens, when the synchronization efficiency is ~ 50%, and there are two clearly defined synchronization patterns. A similar multimode synchronization is called chimeric synchronization (or hopping mode) [13], and the methodology for its estimation is closely related to the methodology described in the current section.

### 2.4 Structure of the oscillatory neural network

In the current study, we review ONN in the form of a one-dimensional chain consisting of $N$ coupled oscillators and examine in details the cases of a short chain consisting of two coupled oscillators ($N$ = 2) and a long chain of $N$ = 100 oscillators. The input parameters of the network are the values of the supply currents of the oscillators, and the output parameters are the synchronization metrics of individual pairs of oscillators. The network structure for two oscillators is presented in Figure 4a, and the network structure for a network of $N$ oscillators is shown in Figure 4b.

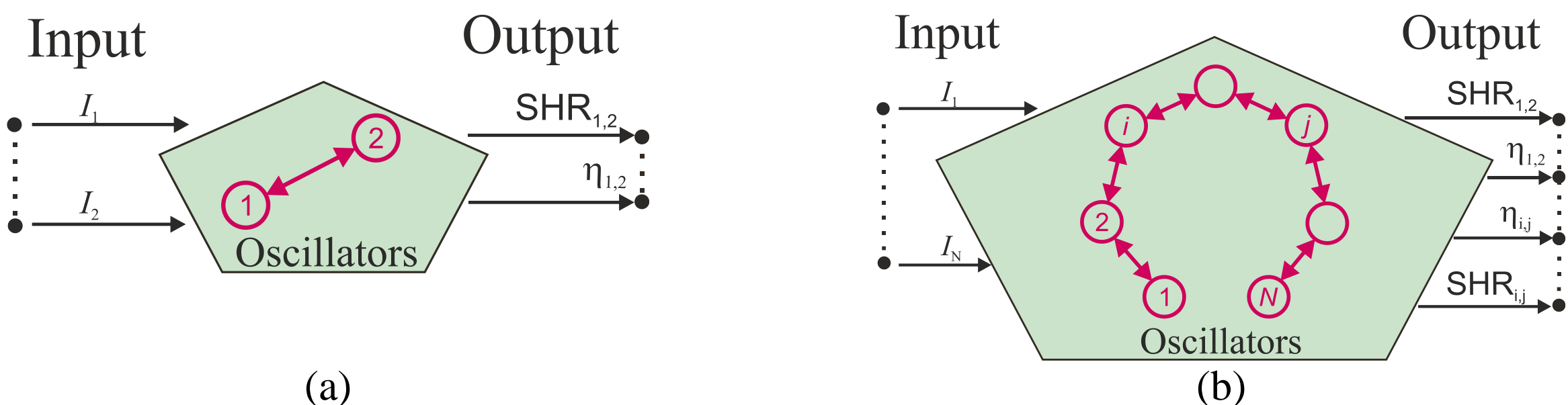

**Figure 4** [Color online] ONN structure with the number of oscillators $N = 2$ (a). ONN structure with the number of oscillators $N$ (b).

## 3. Results

### 3.1 Analysis of experimental results

The dynamics of a system of two thermally coupled planar oscillators (Fig. 1c) was studied by implementing a "strong" coupling ($d$ = 12 μm) and a "weak" coupling ($d$ = 21 μm) (see Methods for details). As the output signal, we used current oscillations taken from the resistor $R_I$. Varying a parameter ($I_1$), which was responsible for the oscillation frequency of the first oscillators, we recorded the oscillation spectra of both oscillators. For the structures with a weak coupling, the current $I_1$ was changing in the range of 660-820 μA, and, in the case of no mutual thermal coupling, the current $I_1$ modified the natural oscillation frequency $F_1^0$ in the range from 1200 to 2000 Hz. The current of the second oscillator remained constant $I_2$ = 720 μA, and the natural frequency was $F_2^0$ ~1550 Hz. Figure 5a demonstrates a set of paired spectra of current oscillations with a step of $\Delta I_1$ = 10 μA.

The spectra demonstrates that in the current range 710 μA ≤ $I_1$ ≤ 750 μA, the oscillators synchronize in frequency, when $F_1 = F_2$, and this effect is observed in the frequency range from 1550 to 1650 Hz.

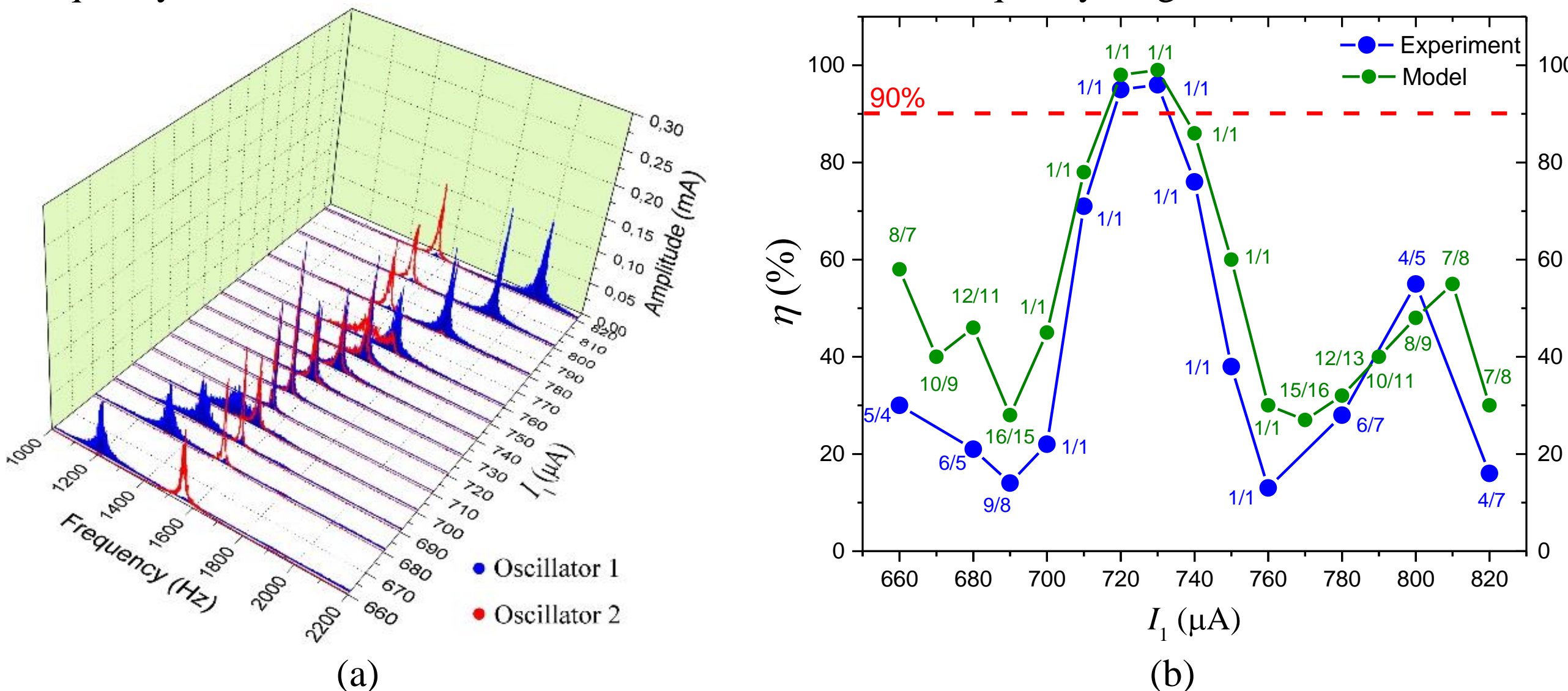

**Figure 5** [Color online] Experimental spectra of current oscillations (a) (Oscillator 1 is a blue graphs and Oscillator 2 is a red graphs), and experimental (blue graph) and model (green graph) curve $\eta(I_1)$ (b) for weak thermal coupling of two oscillators. Model curve $\eta(I_1)$ was obtained at Δ=0.1 V and $U_{n0}$=20 mV. Dashed line marks the level $\eta_{th}$=90%. Fractions denote *SHR* values. $I_2$ = 720 μA.

The algorithm results for the synchronization efficiency $\eta$ and the *SHR* value for experimental oscillograms are presented in Figure 5b (blue graph). The synchronization efficiency varies with the current $I_1$, and, in the range 720 μA < $I_1$ <730 μA, it exceeds the threshold value $\eta_{th}$ = 90%, where the oscillations are considered synchronized. In the synchronization region, the value of *SHR* equals to 1/1, which is consistent with the conclusions drawn from the experimental spectra. The numerical estimates of $\eta$ and *SHR*, calculated using the phase-locking estimation method (Equations 8-11), give a more accurate determination of the synchronization state in the entire current range. A sharp peak in the dependence $\eta(I_1)$, corresponding to *SHR* = 1/1, is observed. The minima $\eta(I_1)$, characterizing the chaotic behavior of the system, is observed at currents $I_1$ = 700 μA and $I_1$ = 760 μA, which correspond to the widest oscillation spectra. A weaker peak is observed at $I_1$ = 800 μA ($\eta$ = 55%), which corresponds to the most probable synchronization *SHR* = 4/5 and indicates the presence of chimeric (multimode) synchronization. However, we attribute these oscillations to the unsynchronized type because $\eta < \eta_{th}$. . Numerical simulation of the oscillograms using Equations (1-4) revealed that the best fit with the experimental spectra of the dependences $\eta(I_1)$ and $SHR(I_1)$ is observed at $\Delta$ = 0.1 V and $U_{n0}$ = 20 mV) (Fig. 5b). Therefore, we can confirm the validity of the thermally coupled oscillator models proposed in section 2.2.

With a distance between the switches $d \sim 12$ μm, under the condition of strong coupling, we observe not only the synchronization at the fundamental harmonics, but also high-order synchronization. Analyzing the shape of the spectrum (Fig. 6a), the type of synchronization can be determined with a confidence only for a few current values. For example, for $I_1$ = 520, the 2nd harmonic of the oscillator 1 coincides with the 1st harmonic of the oscillator 2 and, therefore, *SHR* = 1/2. It is visually easy to determine that in the range 650 μA < $I_1$ <730 μA, synchronization is observed at the fundamental harmonics with *SHR* = 1/1. For a more accurate determination of the distribution of synchronization states, it is necessary to use the phase-locking estimation method of *SHR* calculation (Fig. 6b). The histogram presents the current ranges (blue columns) with clearly defined synchronization, where *SHR* takes a number of discrete values (*SHR* = 1/2, 2/3, 1/1, 4/3, 3/2). The histogram has the shape known as the devil's staircase [14]. The widest region belongs to the value *SHR* = 1/1. Between the regions of strong synchronization, there are regimes of non-synchronized oscillations similar to chaotic oscillations ($\eta < \eta_{th}$, red columns, for example, for *SHR* = 3/4), which exhibit predominantly frequency-blurred spectra.

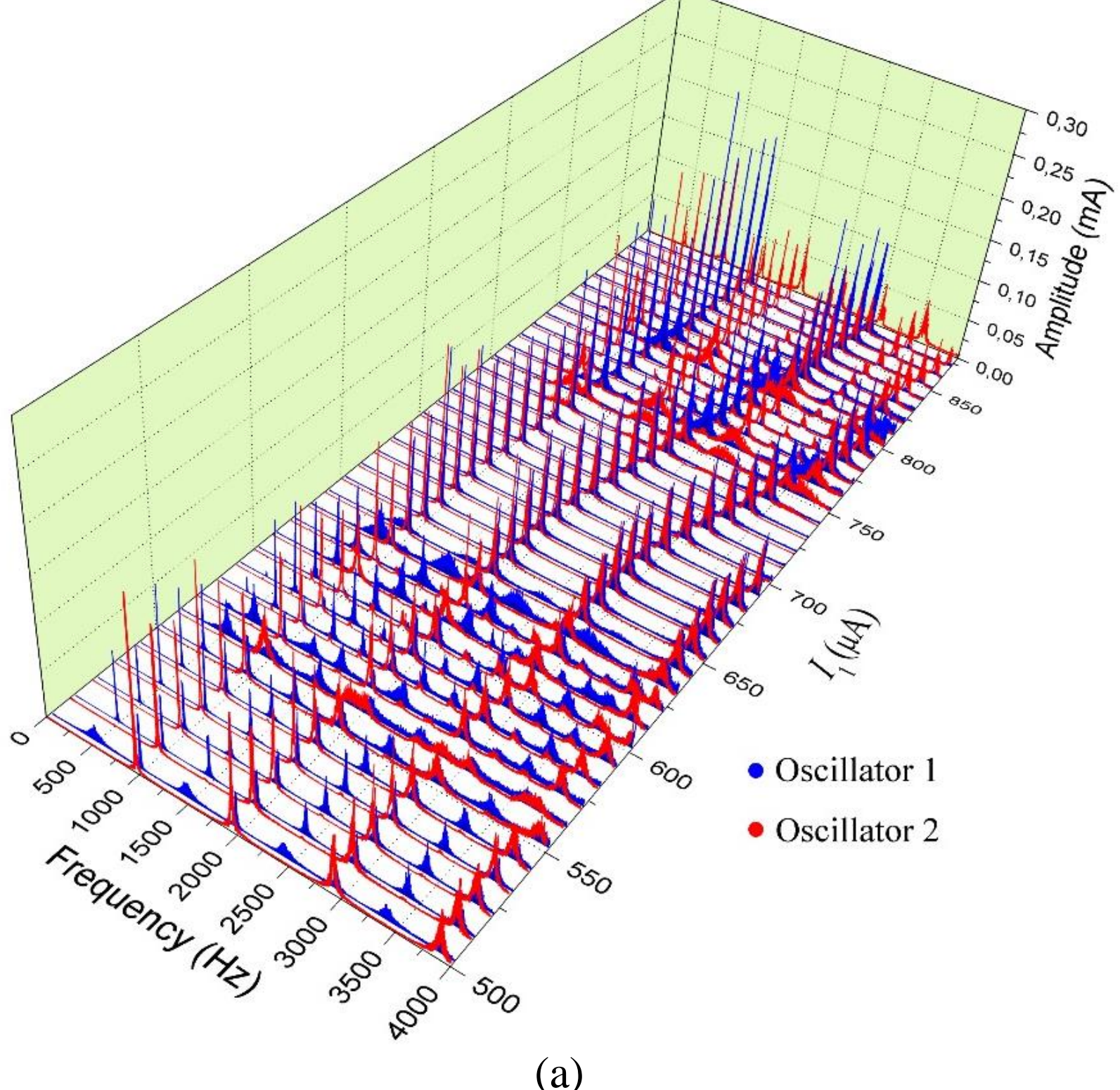

(a)

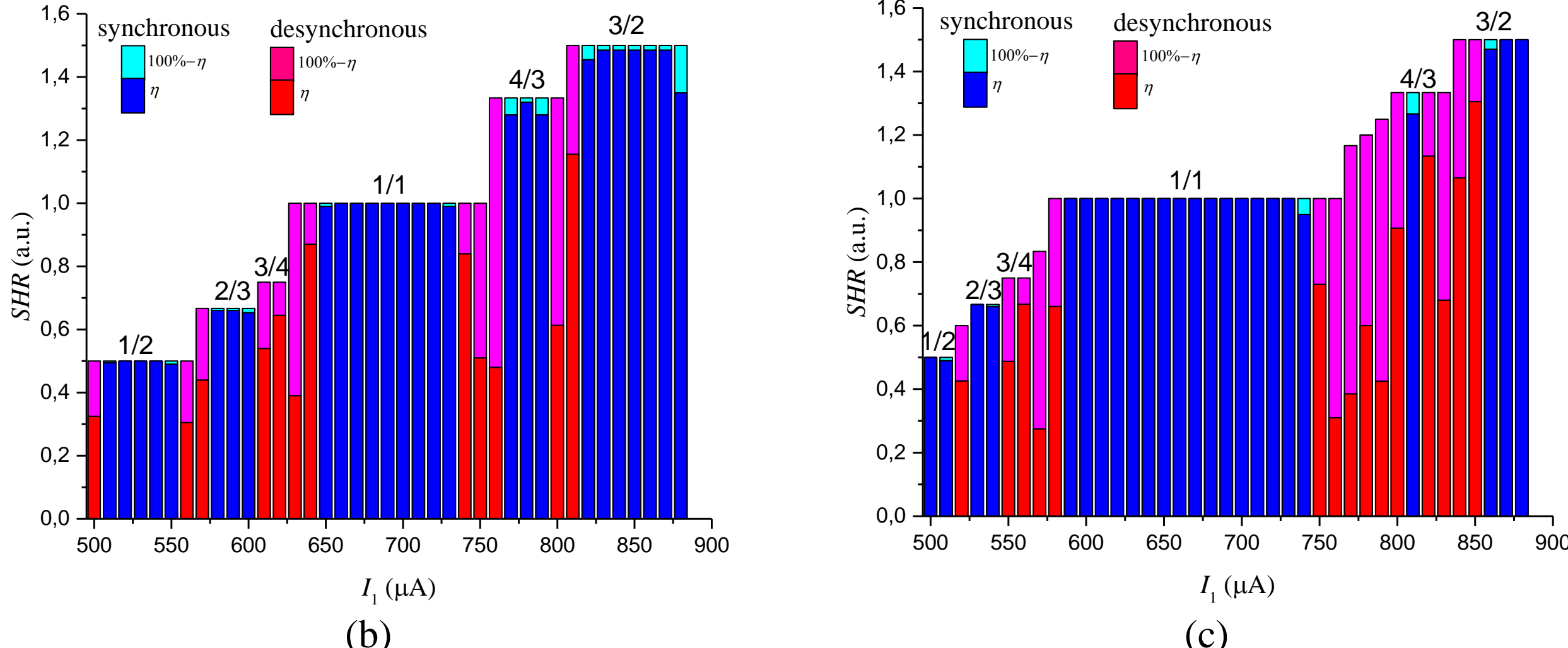

**Figure 6** [Color online] Experimental spectra (Oscillator 1-blue graphs and Oscillator 2-red graphs) as a function of $I_1$ for strong thermal coupling of two oscillators (a). Histograms *SHR* having a specific shape known as the *devil's staircase* [14] and corresponding to experimental (b) and model (c) data. Different colors highlight synchronous ($\eta \geq \eta_{th}$) and desynchronous ($\eta < \eta_{th}$) states and percentage $\eta$ (relative to the height of the corresponding column). $I_2$ = 720 μA.

For the model (Equations 1-4), the parameters were selected ($\Delta$ = 0.5 V and $U_{n0}$ = 20 mV) for the *SHR* distributions to have a histogram qualitatively similar to the experiment histogram (see Fig. 6c). The model histogram has the same set of synchronous states, and a slight difference lies in the current ranges and efficiency values, especially in the areas of chaotic oscillations. The noise level in the model corresponded to the case of weak coupling, but the interaction level, determined by $\Delta$, corresponded to the strong coupling.

Experimentally and by the model, we demonstrated the presence of a high-order synchronization effect in a system of two coupled oscillators. Using the phase-locking estimation method, we accurately estimated the synchronization existence ranges and *SHR* values. The numerical model and the experimentally obtained results were consistent, and the model can be used in the future studies of the high-order synchronization effect.

### 3.2 Research of the system classification capacity

The detection of the set of discrete stable synchronization states $N_s$ in a circuit of only two oscillators can be used for the classification and pattern recognition in a neural network. In the experiment, five states were found ($N_s$=5), however, $N_S$ depends on many parameters and the following sections are devoted to the study of this issue. It is interesting to understand, how *SHR* are distributed in the space of control currents ($I_1$, $I_2$). For parameters corresponding to oscillators with strong coupling ($I_{th_1} < I_1 < I_{h_1}$, $I_{th_2} < I_2 < I_{h_2}$, with a current step of $\Delta I$ = 10 μA, $\Delta$ = 0.5 V and $U_{n0}$ = 20 mV), the model results are presented in Figure 7 (2D and 3D views). The 2D synchronization regions with a specific *SHR* value have the form of Arnold's tongues [14] (colored areas), and these regions are separated by the areas of chaotic behavior of the system (black areas). The 3D distribution type has the form known as the devil's staircase [10]. Many synchronous states *SHR* = (1/3, 2/5, 1/2, 3/5, 2/3, 3/4, 1/1, 4/3, 3/2, 2/1, 5/2, 3/1) consist of $N_s$ = 12 values. The results in Fig. 6c are obtained from the cross section of Fig. 7a at $I_2$ = const = 720 μA.

Here, we introduce the concepts of the effectiveness of the synchronous state $\Psi_{SHR}$ and the level of system synchronism $\Psi$ defined as:

$$\Psi_{\mathrm{SHR}} = \frac{S_{\mathrm{SHR}}}{S} \cdot 100\% \quad , \qquad (12)$$
$$\Psi = \sum_{N_S} \Psi_{\mathrm{SHR}}$$

where $S_{\mathrm{SHR}}$ is the area occupied by a particular synchronous state, and $S$ is the total area of space in the 2D plot (Fig. 7a).

Therefore, $\Psi_{\mathrm{SHR}}$ determines the fraction of a certain synchronous state in the space of control parameters, and $\Psi$ is the sum of the efficiencies of all possible synchronization states and shows the probability of the synchronization occurrence. The larger $\Psi$, the less the probability of non-synchronous oscillations falling into the region with a random choice of $I_1$ and $I_2$. Twelve values of $\Psi_{\mathrm{SHR}}$ in Figure 7a ($\Psi_{1/1}$~20%, $\Psi_{2/1}$~12%, $\Psi_{1/2}$~6%, $\Psi_{3/1}$~3%, $\Psi_{3/2}$~2%, $\Psi_{1/3}$~2%, $\Psi_{4/3}$~1.7%, $\Psi_{2/3}$~1.4%, $\Psi_{3/4}$~1%, $\Psi_{5/2}$~0.36%, $\Psi_{2/5}$~0.3%, $\Psi_{5/3}$~0.14%) show that the highest efficiency occurs during synchronization at the fundamental harmonics *SHR* = 1/1. It is expected due to the symmetry of the oscillators; the overall synchronism of the system is $\Psi$ ~ 50%.

Another question arises, how does the *SHR* spatial distribution and the $N_s$ value depend on the coupling level Δ and the noise value $U_{n0}$? Figures 7b, d present the 2D and 3D *SHR* distributions for the case Δ = 0.3 V and $U_{n0}$ = 2 mV. By changing Δ and $U_{n0}$, it is possible to increase the number of synchronization regions $N_s$ = 82, which are still the linear regions with diagonal symmetry. The $\Psi_{\mathrm{SHR}}$ values of individual states decrease, and the overall synchronism increases $\Psi$ ~ 61%. Figure 8 gives an answer to this question in more details.

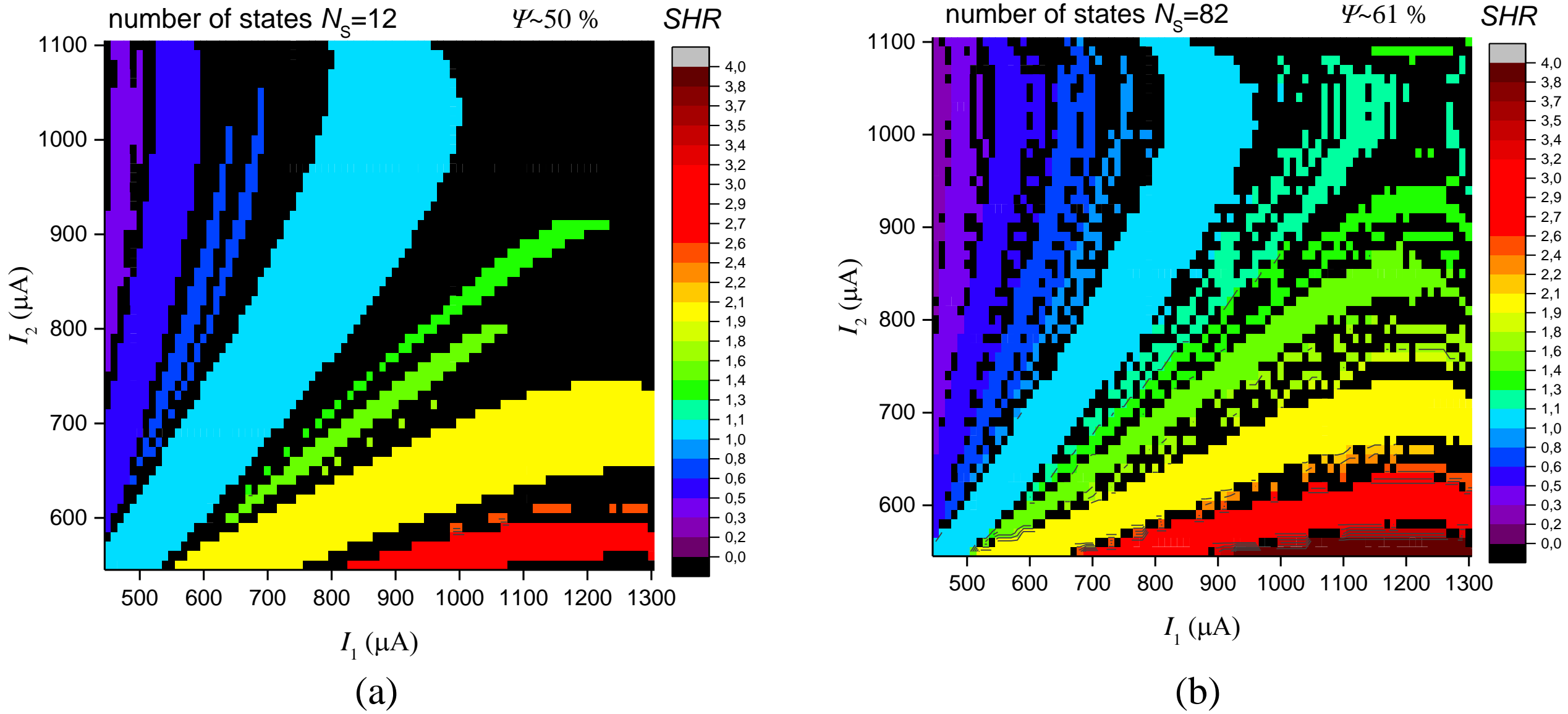

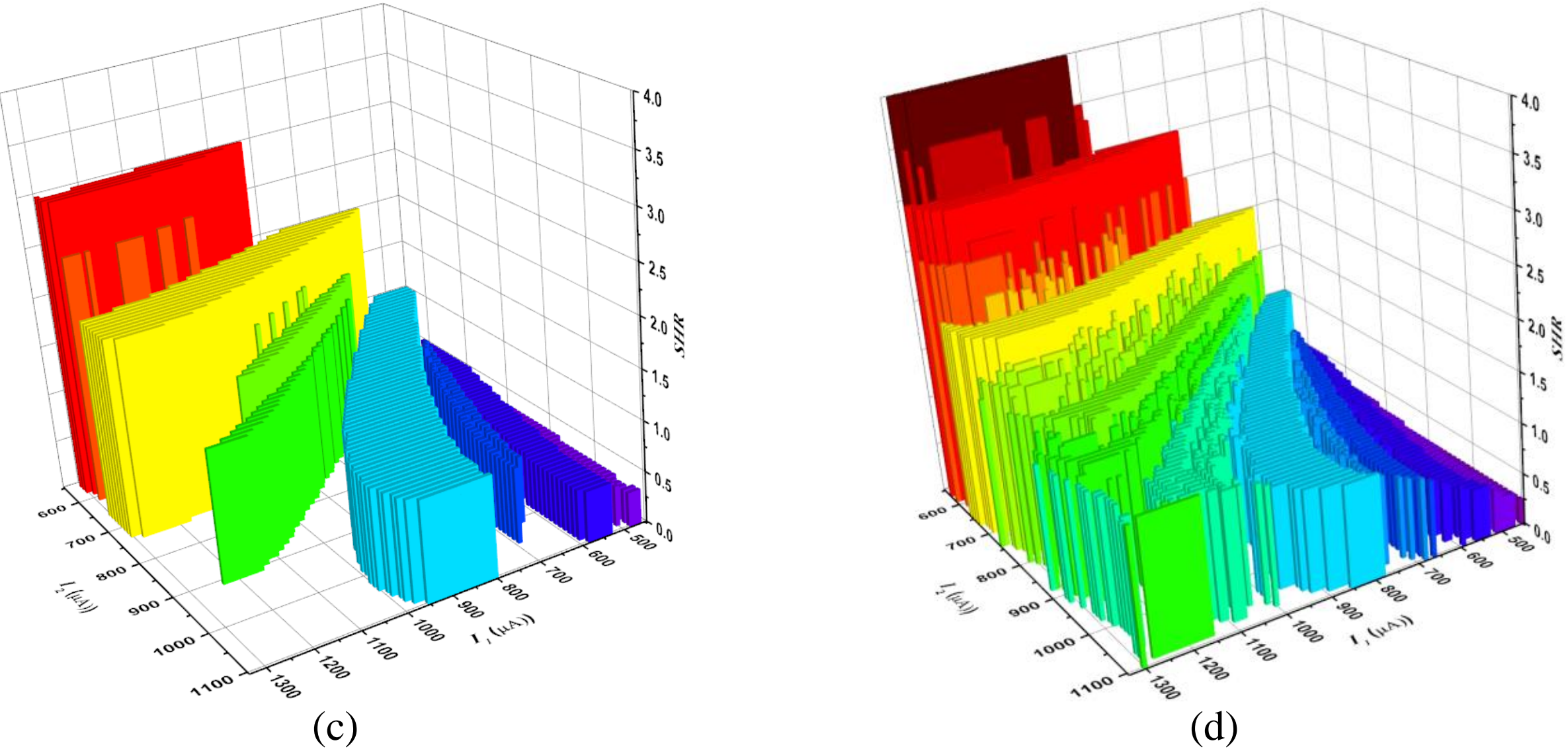

**Figure 7** [Color online] Calculation results of 2D and 3D distributions of *SHR* for model parameters corresponding to experimental oscillators ($\Delta$=0.5 V and $U_{n0}$=20 mV) (a, c) and to oscillators with parameters ($\Delta$=0.3 V and $U_{n0}$=2 mV) (b, d). 2D distribution of *SHR* has the form known as the *Arnold's tongues*, and 3D has the form known as the *devil's staircase*.

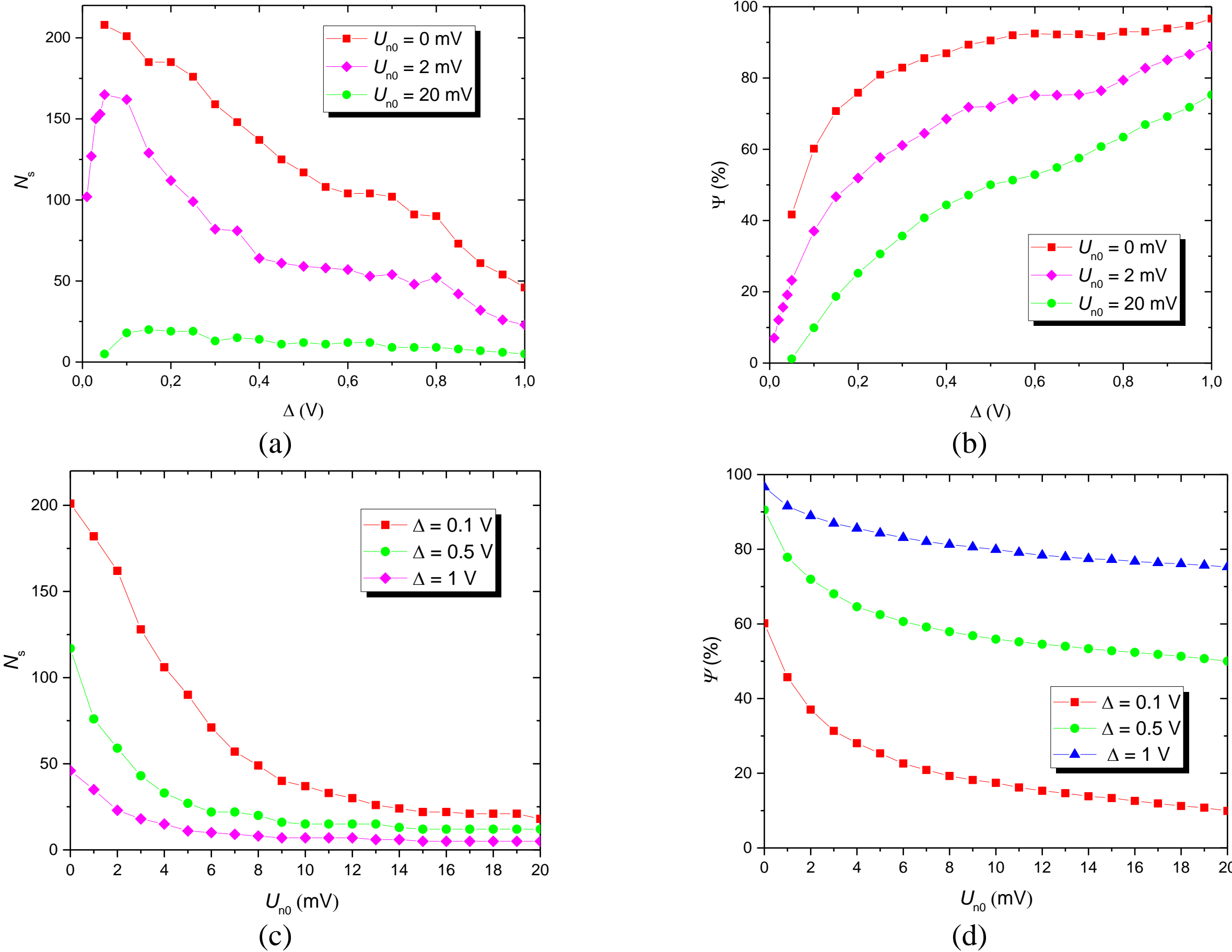

**Figure 8** [Color online] Functions of number of synchronization states $N_s$ (a) and level of system synchronism $\Psi$ (b) on the value of $\Delta$ at fixed values of noise $U_{n0}$; functions of $N_s$ (c) and $\Psi$ (d) on noise $U_{n0}$ at fixed values of $\Delta$.

Figure 8a demonstrates the dependence of the number of synchronization states $N_s$ on the value of thermal coupling $\Delta$ for fixed noise values $U_{n0}$. For the curves where $U_{n0} \neq 0$, it is possible to determine the optimal value of the coupling $\Delta_{opt}$, when a maximum of $N_s$ occurs. $\Delta_{opt}$ shifts to lower values with decreasing $U_{n0}$. The appearance of the maximum, is, most likely, associated with the competition of two factors: the presence of noise, which plays a desynchronizing role, and the presence of coupling, which leads to synchronization. When the coupling level changes from 0 to $\Delta_{opt}$, a sharp increase in the number of synchronization states $N_s$ occurs. A further increase in $\Delta$ causes a decrease in $N_s$, which is associated with the enlargement of synchronization regions due to merger of the regions with low *SHR* order (for example, *SHR* = 1/2) with the regions with higher *SHR* order (e.g. *SHR* = 11/20). A rather high values of $N_s > 150$ at low noise are achieved, and even at the experimental noise level $U_{n0} = 20$ mV, the maximum $N_s$ is about 20 ($\Delta_{opt} = 0.15$ V), which is a high value for a system of only two oscillators. Figure 8b demonstrates the dependence of the level of system synchronism $\Psi$ on the value of the thermal coupling $\Delta$ for various $U_{n0}$. All curves have asymptotic behavior and aim to the level $\Psi = 100\%$, and the system has a larger $\Psi$ for lower $U_{n0}$. The increasing dependence $\Psi(\Delta)$ is associated with the effect of capturing the areas with no synchronization (Arnold's tongues broaden, capturing dark areas, see Fig. 7a, b). In relation to the classification problem, the level of system synchronism reflects the ability of objects classification.

Figure 8c demonstrates the dependence of $N_s$ on the noise value $U_{n0}$ at three levels of coupling force $\Delta$. All curves reflect a decrease in $N_s$ with increasing noise amplitude. With an increase in the noise power, the power attributed to higher-order harmonics becomes insufficient to synchronize the oscillators at high harmonics. The synchronization levels with a high order disappear, and, at maximum noise, synchronization remains only at the fundamental harmonics (*SHR* = 1/1). The absolute value of the derivative $|\partial N_s/\partial U_{n0}|$ decreases with increasing coupling level $\Delta$, it can be interpreted as an increase in noise resistance.

The dependence of the level of system synchronism on the noise value $U_{n0}$ at various $\Delta$ levels is presented in Figure 8d. An increase in $\Delta$ and a decrease in $U_{n0}$ contribute to an increase in $\Psi$. The most significant changes are observed at low noise $U_{n0}$ <4mV, it is, probably, related to a strong change in $N_s$ in this region (see Fig. 8c).

Finding $N_s$ by phase-locking estimation method, we calculated *SHR* values under the condition ($M_1$ and $M_2$) ≤ 20. If we represent the space of possible values of $M_1$ and $M_2$ (Fig. 9), then the maximum number of possible synchronous states defined by Equation (8) is $N_s = 255$; it is equivalent to the number of red cells in the region (20x20). The blue cells in Figure 9 correspond to states that are multiple fractions of the simpler fractions. For example, from the fraction 1/2 we can get 2/4, 3/6, 4/8 fractions, and these states are not taken into calculations. If the absence of synchronization is counted as a separate state, then the total number of states, we call it classification capacity $W_C$, would be $W_{C_20} = 256$.

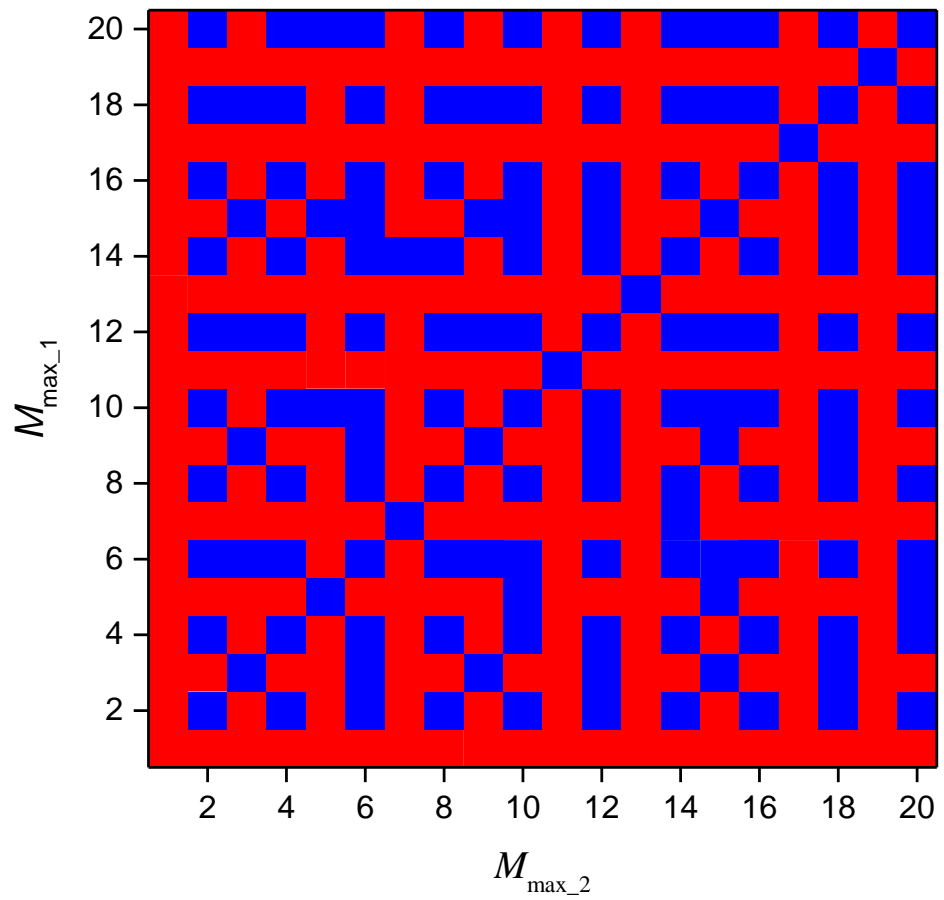

**Figure 9** [Color online] Combinations of the allowed (red squares) and not considered (blue squares) synchronous states, provided that ($M_1$ and $M_2$) ≤ 20. The total number of synchronous states is $N_{s_20} = 255$.

Depending on the chosen limit for $M$, the number of possible states can take the following values ($W_{C_3}$=8, $W_{C_4}$=12, $W_{C_8}$=44, $W_{C_10}$=64, $W_{C_14}$=128, $W_{C_20}$=256, $W_{C_100}$=6088). Here, an analogy with binary logic can be applied, where the capacity of a number depends on the number of bits $n$ and is defined as $2^n$ (1, 2, 4, 8, 16, 32, 64, 128, 256 ...). This can be used in the future for the implementation of the interface between an oscillatory network and a computer.

### 3.3 Technique of storing and recognizing vectors using synchronous states.

A system of $N$ oscillators can store the vectors of dimension $M \leq N$-1 [11] . The coordinate of the stored vector can be expressed by any parameter of the oscillator system that affects the output synchronization, for example, oscillator currents or coupling coefficients $\mathbf{E}$=($I_1$(1), $I_2$(2), $\Delta_{i,j}$(3)…$\Delta_{i,j}$($M$)). We demonstrate the principle of storing and recognizing vectors with the following example.

Let us consider three stages in the ONN of two oscillators (Fig. 4a) that have a synchronization distribution corresponding to the experimental oscillators (Figure 10a).

**I. Before vector storing**

We need to store two vectors with coordinates $\mathbf{E}_1$=(630 μA) and $\mathbf{E}_2$=(1050 μA), where the coordinate is the current value $I_1$ of the first oscillator. Vectors contain only one coordinate $M$ = 1, since N = 2. Before storing, we set the remaining parameters of the system in an arbitrary way, for example, $I_2$=1050 μA, Δ=0.5 V and $U_{n0}$=20 mV. When feeding the vector $\mathbf{E}_1$ and $\mathbf{E}_2$, the oscillators remain in an asynchronous state (see Figure 10a, 'Before storing').

**II. Vector storing**

The process of storing consists in varying the parameters $I_2$, Δ and $U_{n0}$ until both vectors $\mathbf{E}_1$ and $\mathbf{E}_2$ lead the system to synchronization with unique value $SHR_{1,2}$, and the maximum possible value of the synchronization efficiency $\eta$. When selecting parameters, stochastic optimization methods can be used (see Discussion section). In our example, it is enough to set $I_2$=850 μA , leaving Δ and $U_{n0}$ unchanged. This is equivalent to shifting the mappings of vectors on the plane to the synchronization region of each vector (see Figure 10a, 'Vector storing'). As a result, feeding the vector $\mathbf{E}_1$ into the system will lead to synchronization $SHR_{1,2}$=0.75, and the vector $\mathbf{E}_2$ will cause synchronization $SHR_{1,2}$=1.33 with an efficiency of $\eta$ ~ 98%.

**III. Vector recognition**

When unknown vectors $\mathbf{T}_1$ and $\mathbf{T}_2$ are fed into the system (see Figure 10b), the system will take on certain values $SHR_{1,2}$ and $\eta$. For example, the vector $\mathbf{T}_2$ will lead to a lack of synchronization, or to a small value of $\eta$, and it will not be associated with any of the stored vectors. The input of the vector $\mathbf{T}_1$ will lead to synchronization $SHR_{1,2}$=1.33, which associates it with the vector $\mathbf{E}_2$. The degree of match can be estimated by the formula $_m$= 100%-($\eta$($\mathbf{E}_2$)- $\eta$($\mathbf{T}_1$)). The closer the synchronization efficiency to the maximum value, the greater the correspondence of the vectors $\mathbf{E}_2$ and $\mathbf{T}_1$.

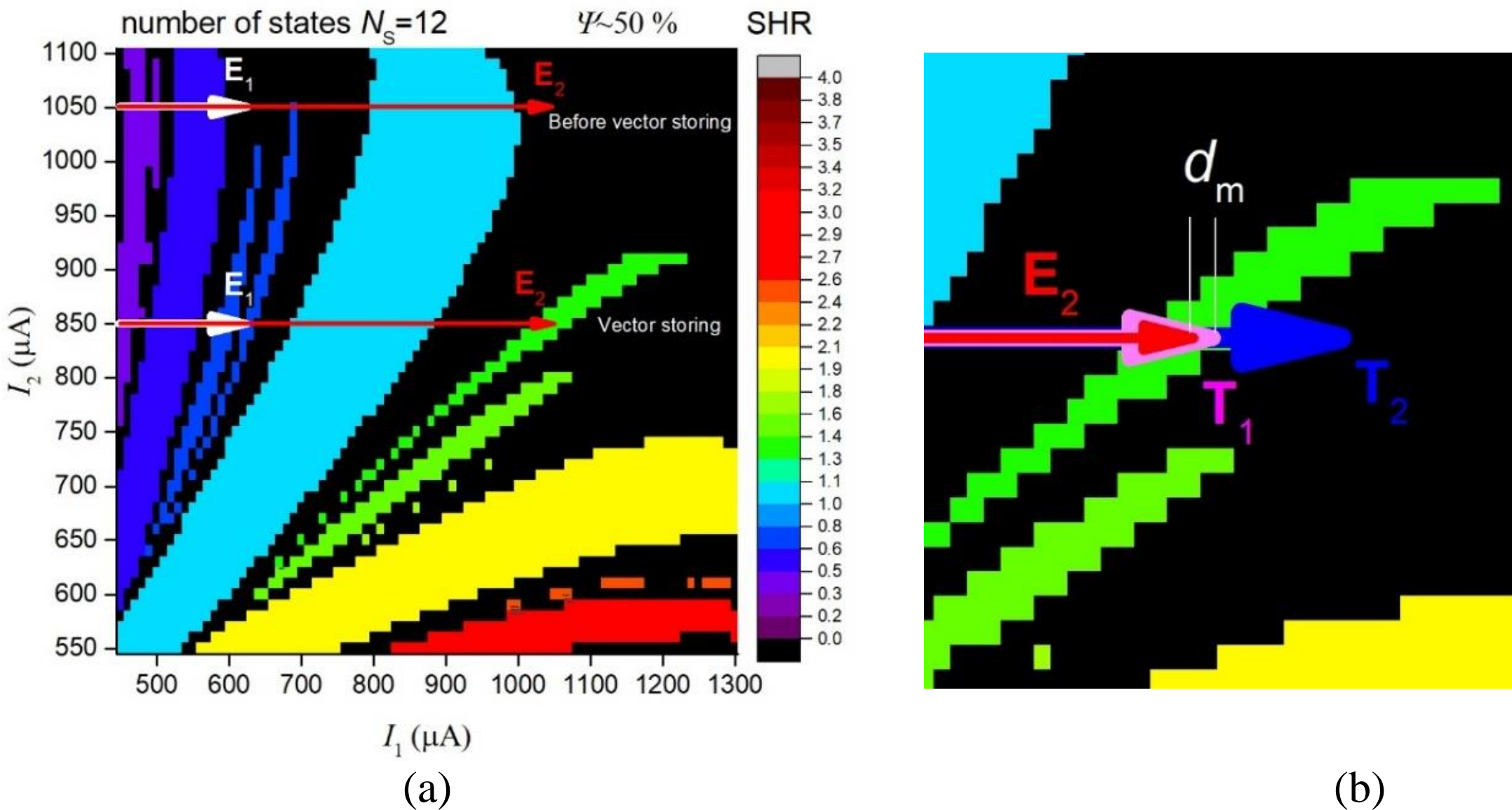

(a) (b)

**Figure 10** [Color online] One-dimensional vectors $\mathbf{E}_1$ and $\mathbf{E}_2$ on the plane of synchronization distribution for the steps 'Before vector storing' and 'Vector storing' (a), 'Vector recognition' (b).

The greater the number of synchronous states $N_s$ the oscillator network can possess, the more vectors this system can store.

The technique's disadvantage is that not all the stored vectors **E** may have a free synchronization region in the desired region of the parameter space. Another issue may arise, when an unknown vector **T** is inputted, and the oscillators synchronize in the areas where the vector **E** has not been stored. As a result, synchronization will take place, but recognition will be undefined.

These issues can be resolved by adding external single-layer neural networks to the system - an input and output linear classifier with $W_{in}$ and $W_{out}$ weights (see Figure 11). The technique turns into a reservoir computing technique [64] with the need to adjust the weights' parameters of the external network.

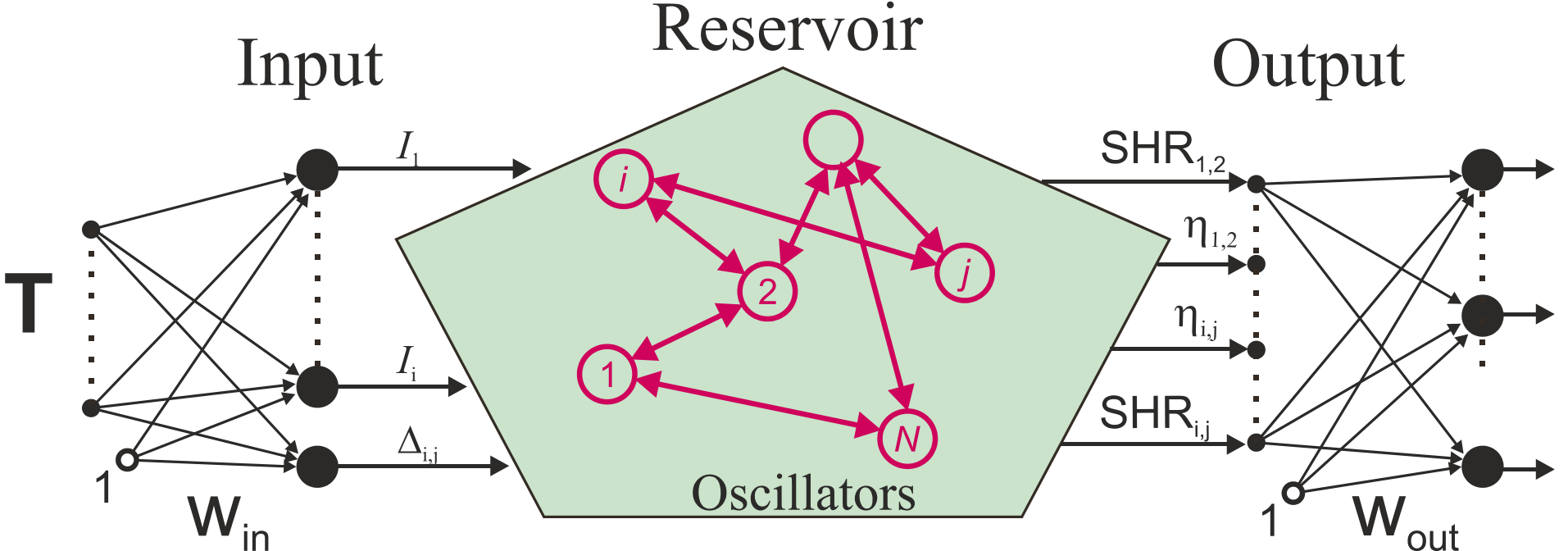

**Figure 11** [Color online] The reservoir computing circuit based on a system of coupled oscillators.

The reservoir can consist of a large number of oscillators and have a significant number of synchronous states $N_s$ . The input vector **T** through the input linear classifier is transformed into a set of current values $I_1$ and coupling coefficients $\Delta_{i,j}$. The output synchronization parameters $SHR_{i,j}$ and $\eta_{i,j}$ from certain pairs of oscillators proceed to the output linear classifier. Therefore, the addition of an external neural network and the corresponding adjustment of the $W_{in}$ and $W_{out}$ coefficients circumvent the physical limitations of the oscillator network and achieve the desired result for storing and recognizing vector images. Efforts towards the development of these techniques will be continued by the authors of the current study.

### 3.4 Effect of a long-range synchronization

The one-dimensional chain model under study is presented in Figure 2b and is described by Equations (1-5). If the synchronization between each neighboring oscillator $SHR_{j,j+1}$ occurs, then the synchronization between the first and the last oscillator of the chain is described by the equation:

$$SHR_{1,N} = SHR_{1,2} \cdot SHR_{2,3} \cdot \ldots \cdot SHR_{N-1,N} \cdot \tag{13}$$

For a system of three oscillators, for certain parameters of the supply currents, we can observe a situation, where synchronization is expressed by the following sequence $SHR_{1,3}$=(1/2)·(3/4)=(3/8). For a chain of five oscillators, the synchronization would be $SHR_{1,5}$=(1/2)·(3/4)·(4/3)·(2/1) =(1/1). Therefore, synchronization on subharmonics allows us to synchronize uttermost oscillators, and to implement long-range synchronization through pairwise synchronization of the oscillators that make up the communication chain. Another discovered effect is the long-range synchronization mode, when the synchronization of neighboring oscillators inside the chain has an efficiency below the threshold ($\eta < \eta_{th}$), and, formally, the oscillators are not synchronized. For example, for a chain of five oscillators, we can observe a situation, where $SHR_{1,5}$=(1/2, $\eta$=97%)·(3/4, $\eta$=50%)·(4/3, $\eta$=67%)·(2/1, $\eta$=92%) =(1/1, $\eta$=97%). In brackets, we indicate the synchronization efficiency of each pair of oscillators. Therefore, the synchronization on subharmonics allows the efficient transmission of synchronization even with weak synchronization of its individual links. The reason for this effect is, apparently, the coexistence of several types of synchronization patterns in the signal (chimeric synchronization [13]), which transmit the interaction, even the signal efficiency is below the threshold. The synchronization effect in the chain of oscillators requires a detailed study and can be used, for example, to create logic elements for computational problems.

We studied the long-range synchronization in ONN with a long chains of oscillators (Fig. 4b) in the range $N$ = 10-100, and the results were independent of $N$. Results for $N$ = 100 are shown in Figure 12. The synchronization was determined between the 1st oscillator and the 100th oscillator $SHR_{1,100}$. While the currents varied for the oscillators 1 and 2 (Fig. 12a) and for the central oscillators 51 and 52 (Fig. 12b), the currents for all other oscillators $I$ = 750 μA and the coupling levels Δ = 0.3 V between them remained constant.

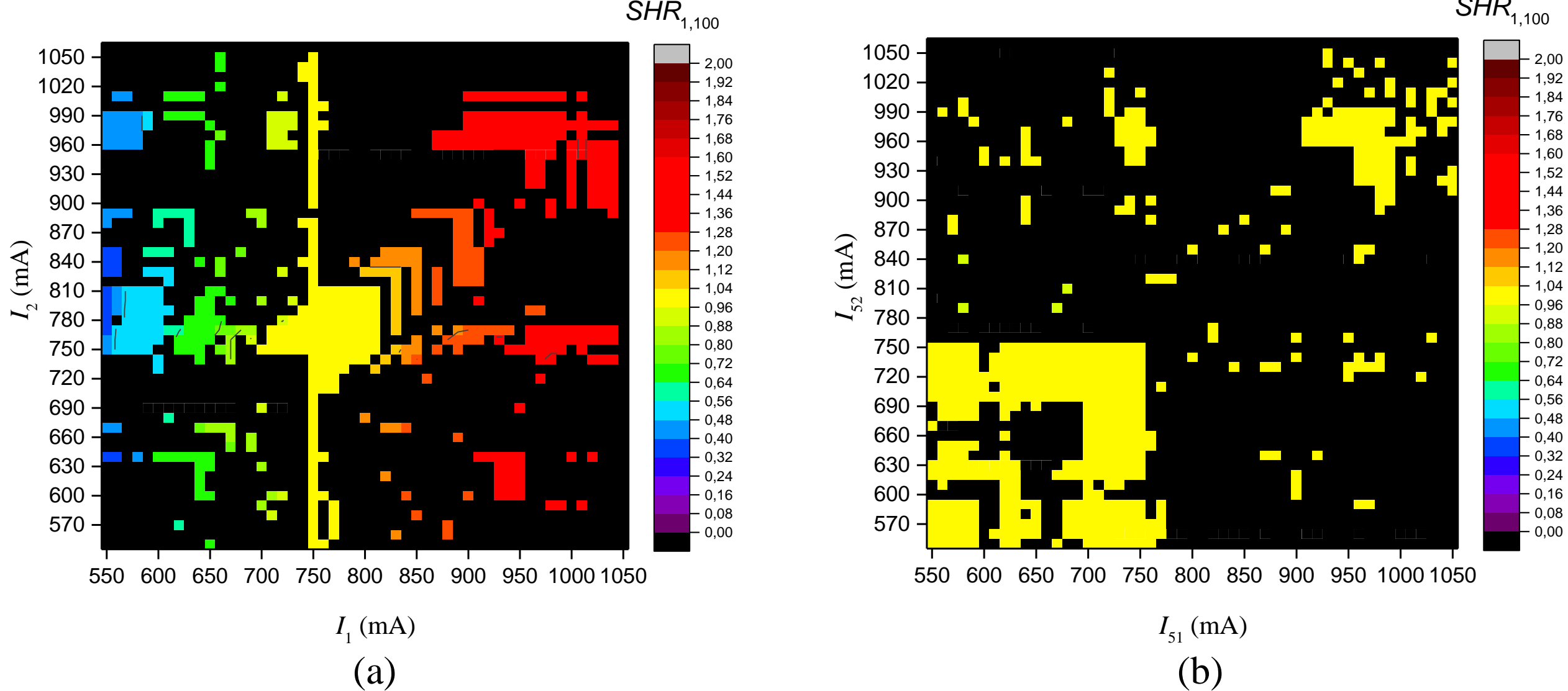

**Figure 12** [Color online] Distribution $SHR_{1,100}$ for a long chain of oscillators ($N$=100) at current variation for oscillators 1 and 2 (a) and oscillators 51 and 52 (b).

According to Figure 12a, the $SHR_{1,100}$ distribution has a change gradient only along the $I_1$ axis, in contrast to the two-oscillator circuit (Fig. 7), because the synchronization was measured between the variable 1st oscillator and constant 100th oscillator. The symmetry of synchronous states remained diagonal, and it indicates the role of the mutual ratio of the currents of neighboring oscillators in the mechanism of synchronism propagation along a chain. However, the synchronization regions are not uniformly distributed and do not form a clear Arnold's tongues pattern.

Figure 12b reflects the distribution of one single state $SHR_{1,100}$ = 1/1, caused by the constant frequencies of the uttermost oscillators. The ratio of $I_{51}$ and $I_{52}$ acts as a key that turns synchronization on or off, and the on-state map is reflected in Figure 12b. Predominantly diagonal symmetry is caused by the equal role of variable oscillators in the propagation of synchronism along the chain.

### 3.5 Application of high order synchronization for computer calculations

In addition to classification tasks, the high-order synchronization effect can be used to create computing oscillatory systems. Here, we outline several approaches. First, Equation (13) demonstrates on a chain of oscillators the implementation of the "multiplication" function. This approach is similar to an analog computing system, as it uses the conversion of the physical parameter *SHR*. Second, a digital binary logic can be used, where the states of a logical "1" or logical "0" can be represented as different types of mutual synchronization of oscillators. For example, if we have two main oscillators in the system (see Fig. 13) designated as oscillator 1 and oscillator 2, then the oscillators are sources of signals of two main levels ("0", "1").

At least three principles can be proposed for dividing signals into logical levels, expressed by the following Equations (14-16).

$$\text{Logic level} = \begin{cases} \text{"1", synchronized with oscillator 1 } (\eta \geq \eta_{\text{th}}) \\ \text{"0", not synchronized with oscillator 1 } (\eta < \eta_{th}) \end{cases} . \quad (14)$$

According to first principle (Equation 14), in order to obtain a logical "1", it is necessary to read the signal directly from oscillator 1, as it is synchronized with itself, and to read a logical "0" from unsynchronized oscillator 2 (when $\eta < \eta_{\text{th}}$). When, for example, passing the "NOT" element, an unsynchronized signal should become synchronized. This effect becomes possible, as synchronization can be transmitted and transformed from the non-synchronous type to the synchronous type due to chimeric synchronization.

$$\text{Logic level} = \begin{cases} \text{"1", } SHR_{1,\text{N}} \text{ is finite decimal fraction} \\ \text{"0", } SHR_{1,\text{N}} \text{ is infinite decimal fraction} \end{cases} . \quad (15)$$

In Equation (15), the degree of synchronization $SHR_{1,\text{N}}$ is measured between oscillator 1 and any other oscillator *N* (see Fig. 13). All possible *SHR* combinations, represented by a fraction, can be divided into two classes: infinite fractions and finite decimal fractions. Figure 13 demonstrates two "NOT" elements. The signal with synchronization $SHR_{1,2}$ = 1/3 = 0.333(3), which is an infinite decimal fraction, is sent to the input of the upper element, and the signal $SHR_{1,\text{N}}$ = 1/4 = 0.25, which is the finite fraction, appears at the output. The signal taken directly from the oscillator 1 is always expressed by the finite fraction, because $SHR_{1,1}$ equals to 1/1 = 1 and corresponds to the logical "1".

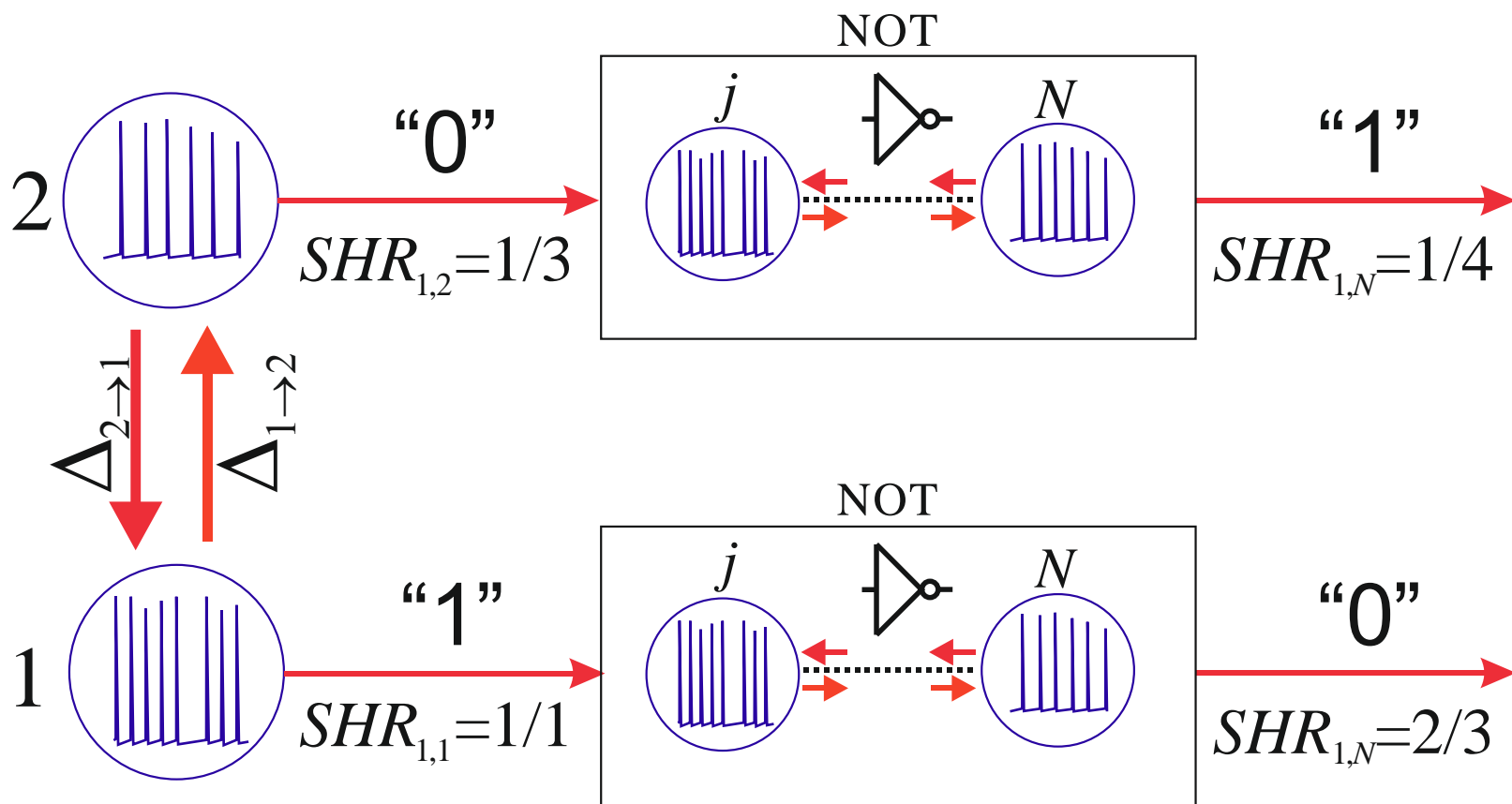

**Figure 13** [Color online] Principle circuit of logical signals sources ("1" and "0"), formed by coupled oscillators 1 and 2 and by two logical elements "NOT", functioning according to Equation (15).

The effect of conversion from a finite fraction to an infinite fraction can be expressed by Equation (13), for example, $SHR_{1,N}$=(1/3)·(3/4)=(1/4).

The third principle is to compare the *SHR* value with a certain threshold value, for example, by the formula:

$$\text{Logic level} = \begin{cases} \text{"1"},\ SHR_{1,\text{N}} \geq 1 \\ \text{"0"},\ SHR_{1,\text{N}} < 1 \end{cases}. \tag{16}$$

On order to implement the logic element, it is necessary to select the parameters of the oscillators constituting the element in a way that the synchronization transfer coefficient varies depending on the logical level of the input signal. The development of such circuits can be the subject of a future study.

We have outlined several options for calculations based on the high-order synchronization effect. Some methods (Equations 13-16) can be implemented only for describing oscillations with the fractional *SHR* parameter, and these methods are not suitable for the traditional synchronization effect at the fundamental harmonics.

## 4. Discussion and conclusion

We studied the circuit of two thermally coupled $VO_2$ oscillators that exhibits a high-order synchronization effect, which can be used in the methods for classification and pattern recognition in an oscillatory neural network. The experimental circuit, consisting only of two oscillators, has a number synchronous of states $N_s$ = 12. In addition, we demonstrated by modelling that at certain levels of coupling Δ and noise $U_{n0}$ much higher values of $N_s$ > 150 can be achieved. Compared to Vodenicarevic et al. [18], where the circuit consisted of 4 oscillators, had a maximum $N_s$ = 9 and was based the classical effect of synchronization at the first harmonics, our new method for isolating multi-stable states of the system through synchronization on subharmonics has a clear advantage.

We use our phase-locking estimation method for determining the subharmonic ratio and synchronization efficiency for spiking oscillations. The method has several advantages and solves the *SHR* search problem more efficiently than the spectral approach.

The maximum theoretical classification capacity $W_C$ of two oscillators can vary widely depending on the selected parameters of the phase-locking estimation method. The capacity can take values such as $W_C$ = (8, 64, 128, 256), and it leads to the idea of creating an interface between oscillatory and binary computer logic. For example, if the input information presented in the form of a single byte takes one of 256 values, then this information can be entered into the oscillator network by the corresponding synchronous state of the input oscillator in the range $W_{C_20}$ = 256 of different *SHR* values. After processing the information by the neural network, the *SHR* value on the output oscillator is mapped to the value of the output byte. In this way, the input and output information of the oscillatory neural network can be represented in digital form, and SNN can be interfaced with computer interfaces.

At low noise, $W_C$ can take even higher values (for example, $W_{C_100}$ = 6088), and the input information to one oscillator can be represented not by a single byte but by several bytes.

We introduced the concepts of the synchronous state efficiency $\Psi_{SHR}$ and the level of synchronism $\Psi$, which determine the probabilities of synchronization with a certain *SHR* value and the probability of synchronization of the system as a whole. It allow us to assess the suitability of the system for solving practical problems.

In this study, we presented a thermal coupling model of two oscillators, which demonstrated a good fit in the test with the experimental data. By modelling the circuit, we identified the dependences of the number of synchronization states on the coupling value and noise level. Results indicated that the dependence $N_s$ ($\Delta$) has a maximum, and an increase in noise always leads to a decrease in $N_s$ and $\Psi$. The *SHR* distributions in the area of control parameters have the form of Arnold's tongues, and present the linear regions with diagonal symmetry.

The modelling demonstrated that the main parameters affecting the $N_s$ value are the noise level in the system and the coupling value of the oscillators. The adjustment of the thermal coupling in the experiment can be done by changing the distances between the structures. For resistive and capacitive couplings [51], it is sufficient to change the value of the coupling elements. The minimum noise level in the oscillatory network depends on the physical properties of the used switching elements and external interference. The noise level can be increased by adding a noise generator to the system, but to reduce the minimum possible noise value is more difficult task. In our experiment, the minimum attainable noise level was determined by the noise of resistive $VO_2$ switches ($U_{n0}$ = 20 mV), which leads to the instability of the output frequency with a relative period jitter of ~ 0.4%. With this noise level, we reached $N_s$ = 12, and in the case of optimal coupling, we can achieve $N_s$ = 20 (see Figure 8a). The study of noise in $VO_2$ switches and period jitter of oscillations is a separate experimental task. To reduce noise in the relaxation oscillator, the low-noise switches, for example, based on semiconductor thyristors, or low-power RC relaxation oscillators with 0.015% relative period jitter can be used [65]. Therefore, the model data regarding $N_s$ values presented in current study can be achieved in the experiment.

Chapter 3.3 outlines the technique for storing and recognizing vectors, where the storage capacity depends on the value of $N_s$. The process of vector storage is, in fact, the training of a neural network using random selection of parameters in a given interval, similar to the model annealing method. Stochastic approaches to training (optimization) are often used in recurrent networks with feedbacks, as the back propagation method of error is not effective in this case [66]. An example of the stochastic approach to training an oscillator network is presented in our previous work [12]. The method requires large computational resources and time, and the optimization of ONN training methods is a relevant task for future research. An oscillator network is a type of recurrent neural network (RNN) with feedback. The complexity and inefficiency of existing algorithms for RNN training call for new approaches and strategies to utilize computational capabilities of RNN. One of the new approaches is reservoir computing (RC). The structure of the reservoir network is presented in Figure 11. RC uses RNN as a reservoir with rich dynamics and powerful computing capabilities. The reservoir is formed randomly, and it eliminates the need of reservoir training. Instead, only the coefficients of the reading layer are trained, which is responsible for the required task of classification, prediction, and clustering. The main task of the reservoir is to translate the input data into a higher dimensionality of space so that the output neural network can classify the result more correctly. A similar idea for machine learning is called a kernel trick [67,68]. Earlier we demonstrated that the output parameters of the ONN-based reservoir can be $SHR_{i,j}$ and $\eta_{i,j}$ synchronization metrics, however this is not the only way to represent the output data. For example, the ONN output can be presented as the phase difference between the signals of individual oscillators [17,69]. The phase difference can be discrete [69] or continuous [17]. The detection of phase difference requires a large number of phase detectors, and it significantly reduces the applicability of this solution. Another way of presenting information to the ONN is frequency coding [17], when the input data is encoded by the frequency offset of the oscillators relative to some fixed frequency, and the output data is encoded by the average amplitude of all signals. In some cases, the output data is directly encoded by the output frequencies [18,70]. These methods often require computational overhead for the Fourier transform.

The effect of long-range synchronization is demonstrated, where the synchronization of uttermost oscillators is expressed in terms of the product of *SHR* values of the intermediate links. In addition, the remarkable property of the oscillator chains is demonstrated to transmit synchronization even for small values of the synchronization efficiency of intermediate pairs of oscillators. In our opinion, this effect is caused by insufficiently studied effect of chimeric synchronization, described in [13]. The stability of the transmission mechanism of high-order synchronization over long distances is demonstrated, and it opens promising opportunities for practical applications. The synchronization distribution does not form a vivid picture of Arnold's tongues, but individual oscillator pairs play the role of a key that turns on or off the synchronization of uttermost oscillators. The presented long-range synchronization effect allows connecting a large number of oscillators influencing each other in a single network through the operation of "multiplication", and it increases the classification capacity of the neural system. The storage capacity can be increased by adding an input and output linear classifier to the system, by analogy with the systems for reservoir computing [64].

In addition, we presented the principles of organizing computations based on the high-order synchronization effect, in particular, the analog principle of multiplication and the digital approach for implementing binary logic. We implemented the principle of dividing a signal by a logical "1" and logical "0" using a synchronous and non-synchronous state to create a neural network information converter [13] capable of performing image conversion, classification, and filtering operations. The principles for organizing the separation of levels have not yet found practical implementation, and can present the subject for future research.

The study results demonstrate the possibility of creating thermally coupled oscillatory structures based on $VO_2$ elements with an S-shaped current-voltage characteristic, and it opens the way to creating a network with a large number of oscillators and a large variety of functions. The technological possibility of developing arrays of $VO_2$ switches on a chip is available, as $VO_2$ sensors are used to manufacture bolometric matrices [71]. Along with $VO_2$ switches, other switching elements with an S-shaped I–V characteristic can be used in the presented circuits, for example, based on $NbO_2$ [72] or chalcogenide materials [73]. The manufacturing technologies for these alternative materials are already developed and used to create elements of non-volatile memory from Intel [74].

The effects of increasing the classification capacity of oscillator circuits described in this study and the principles of calculation based on the universal physics effect of higher order synchronizations can be applied to most oscillators with any type of coupling. It increases the practical significance of the presented results for developing the capabilities of spiking neural networks and the creation of devices with artificial intelligence.

## Acknowledgements

This research was supported by Russian Science Foundation (grant no. 16-19-00135).
The authors express their gratitude to Dr. Andrei Rikkiev for the valuable comments in the course of the article translation and revision.